\documentclass[10pt,twocolumn,letterpaper]{article}

\usepackage{iccv}
\usepackage{times}
\usepackage{epsfig}
\usepackage{graphicx}
\usepackage{amsmath}
\DeclareMathOperator*{\argmin}{argmin}
\usepackage{amssymb}
\usepackage{wrapfig}
\usepackage{sidecap}
\usepackage{subcaption}
\usepackage{multirow}
\usepackage{makecell}
\usepackage[symbol]{footmisc}
\usepackage{bbm}

\usepackage[pagebackref=true,breaklinks=true,letterpaper=true,colorlinks,bookmarks=false]{hyperref}

\iccvfinalcopy 

\def\iccvPaperID{5051} 

\newcommand{\myparagraph}[1]{\noindent{\bf #1}}

\newcommand{\p}[1]{\textcolor{green}{#1}}

\ificcvfinal\fi
\begin{document}

\title{Monocular 3D Human Pose Estimation by Generation and Ordinal Ranking}

\author{Saurabh Sharma\footnotemark[1] $^{1}$ \hspace{4mm}
        Pavan Teja Varigonda$^{2,3}$ \hspace{4mm}
        Prashast Bindal$^{2}$ \hspace{4mm}
        Abhishek Sharma$^{3}$ \hspace{4mm}
        Arjun Jain$^{2,3}$ \vspace{4mm} \\ 
        \begin{tabular}{ccc}
            $^{1}$Max Planck Institute for Informatics & $^{2}$Indian Institute of Technology &$^{3}$Axogyan AI\\Saarbr{\"u}cken&Bombay &Bangalore
        \end{tabular}
}
\twocolumn[{
\renewcommand\twocolumn[1]{#1}
\maketitle
\begin{center}
    \centering
    \includegraphics[width=\linewidth]{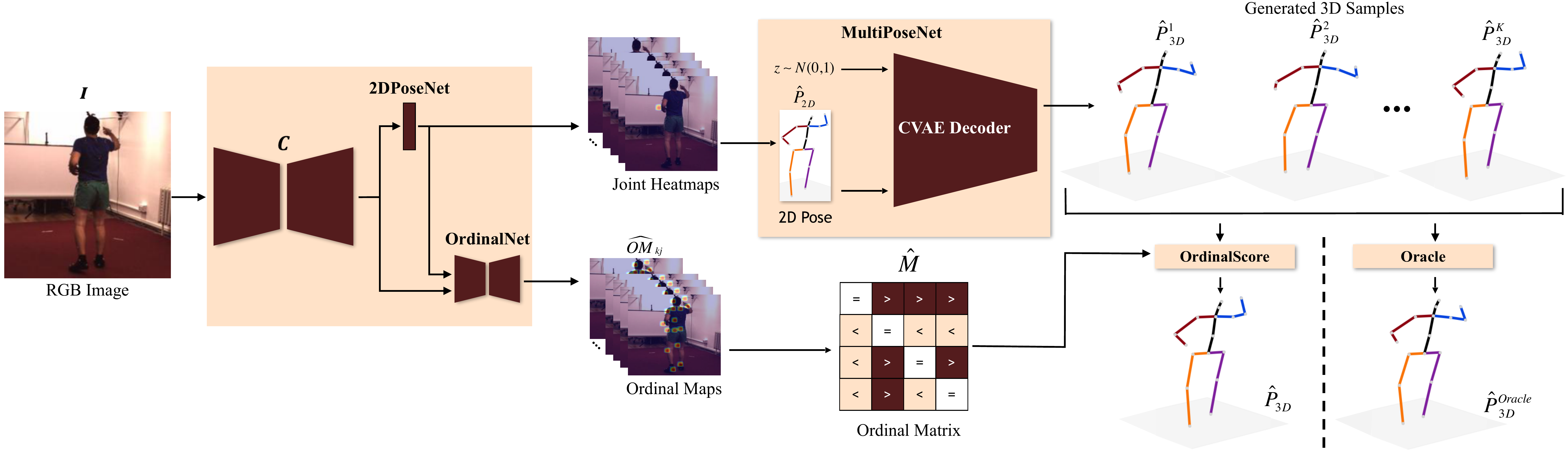} 
     \vspace{-2em}
    \captionof{figure}{A schematic illustration of our modular framework. We tackle the ambiguity in the 2D-to-3D mapping by training a CVAE to generate 3D-pose samples conditioned on the 2D-pose, that are scored and weighted-averaged using joint-ordinal relations, which are regressed together with the 2D-pose. Our upper-bound performance is obtained by using an Oracle. }
     \label{fig:teaser}
\end{center}
}]
\ificcvfinal\thispagestyle{empty}\fi
\begin{abstract}
\vspace{-0.5em}
  Monocular 3D human-pose estimation from static images is a challenging problem, due to the curse of dimensionality and the ill-posed nature of lifting 2D-to-3D. In this paper, we propose a Deep Conditional Variational Autoencoder based model that synthesizes diverse anatomically plausible 3D-pose samples conditioned on the estimated 2D-pose. We show that CVAE-based 3D-pose sample set is consistent with the 2D-pose and helps tackling the inherent ambiguity in 2D-to-3D lifting. We propose two strategies for obtaining the final 3D pose- (a) depth-ordering/ordinal relations to score and weight-average the candidate 3D-poses, referred to as OrdinalScore, and (b) with supervision from an Oracle. We report close to state-of-the-art results on two benchmark datasets using OrdinalScore, and state-of-the-art results using the Oracle. We also show that our pipeline yields competitive results without paired image-to-3D annotations. The training and evaluation code is available at \url{https://github.com/ssfootball04/generative_pose}.
   \vspace{-2em}
\end{abstract}



\section{Introduction}
 Accurate 3D human-pose estimation from a monocular RGB image finds applications to robotics, virtual/augmented reality, surveillance, and human computer interaction. The diverse variations in background, clothing, pose, occlusions, illumination, and camera parameters in real-world scenarios makes it a challenging problem.  The popular 3D-pose annotated datasets do not cover these variations appropriately. Recent advancements in real-world 2D-pose estimation \cite{NewellYD16, wei2016convolutional} has led to several multi-stage architectures, where the 3D-pose is \emph{regressed} either from both the image features and an intermediate 2D representation \cite{bogo2016keep, dabral2018learning, omran2018neural, Zhou_2017_ICCV}, or only the estimated 2D-pose \cite{akhter2015pose,martinez2017simple, Moreno-Noguer_2017_CVPR, varunECCV2012, zhou2016sparseness}.\footnotetext[1]{Majority of this work was done while author was at Indian Institute of Technology, Bombay.} Unfortunately, regression based approaches using only the estimated 2D-pose, ignore the ambiguity in lifting 2D human-pose to 3D: \emph{an inherently ill-posed problem}. Motivated by this shortcoming, we propose to learn a generative 3D-pose model conditioned on the corresponding 2D-pose that affords sampling diverse samples from the learnt 3D-pose distribution. To the best of our knowledge, we are the first to employ a Deep Conditional Variational Autoencoder \cite{sohn2015learning} (CVAE for short) for 2D-to-3D generative human-pose modeling and demonstrate its advantages over direct regression based approaches. We also show that our generative 2D-to-3D module can be trained on a separate MoCap dataset that doesn't have any intersection with the evaluation image-to-3D dataset, and still performs reasonably well.
 Therefore, our modular approach tackles the infeasibility (or high cost) of obtaining 3D-pose annotation for images in real-world and works well with separately collected 2D-pose annotations of real-world images and indoor motion capture data \cite{yasin2016dual}.

Our pipeline is depicted in Figure~\ref{fig:teaser}. First, the \emph{2DPoseNet} head of a deep convolutional network backbone, $C$, estimates 2D pose, $\hat{P}_{2D}$, from a monocular RGB image, $I$. The estimated 2D pose, $\hat{P}_{2D}$, and a latent code $z$, sampled from a prior distribution $p(z)\sim\mathcal{N}(0,1)$, are fed to the decoder of the \emph{MultiPoseNet} CVAE to sample a 3D pose, $\hat{P}^k_{3D}$. Multiple samples, $z^k \in \{z^1, z^2 \dots z^{K}\}$, from $p(z)$ yield a diverse set of 3D pose samples, $\mathcal{S} = \{\hat{P}^k_{3D}: k \in \{1,2,\ldots K\}\}$, consistent with $\hat{P}_{2D}$. Then we employ pairwise depth ordering of body-joints encoded in the estimated joint-ordinal relation matrix, $\hat{M}$, obtained from the \emph{OrdinalNet} head of $C$, to obtain scores, $\{f(\hat{P}^k_{3D}):k \in\{1,2 \dots K\}\}$ for the elements of $S$. These scores are finally fed to Softmax operator to obtain a probability distribution over $S$, reflecting the consistency of the 3D-pose samples to the predicted ordinal relations. The final 3D pose, $\hat{P}_{3D}$, is computed as the expectation of this distribution. Moreover, in order to estimate the upper-bound performance of our generative model, we also report the accuracy w.r.t. the sample, $\hat{P}^{Oracle}_{3D}$, that is the closest match to the ground truth 3D-pose, $P_{3D}$. The \emph{Oracle} upper-bound outperforms all existing state-of-the-art methods, without leveraging recently introduced ordinal dataset, temporal information, or end-to-end training of the multi-stage architectures. This observation supports the strength of our CVAE-based generative model for 2D-to-3D lifting.

A summary of our contributions is as follows - 
\begin{itemize}
    \item We tackle the inherent ill-posed problem of lifting 2D-to-3D human-pose by learning a deep generative model that synthesizes diverse 3D-pose samples conditioned on the estimated 2D-pose.
    \vspace {-0.75em}
    \item We employ CVAE for 3D human-pose estimation for the first time.
    \vspace {-0.75em}
    \item We derive joint-ordinal depth relations from an RGB image and employ them to rank 3D-pose samples.
    \vspace {-0.75em}
    \item We show that the oracle-based pose sample obtained from our proposed generative model achieves state-of-the-art results on two benchmark datasets, Human3.6M \cite{h36m_pami} and Human-Eva \cite{sigal2010humaneva}.
    \vspace {-0.75em}
    \item We show competitive performance over \emph{Baseline} even when our 2D-to-3D module is trained on a separate MoCap dataset \emph{with no images}.
\end{itemize}

\section{Related Work}
\myparagraph{Lifting 2D to 3D} Our approach belongs to the large body of work that obtains 3D-pose from estimated 2D-pose. In \cite{varunECCV2012}, a set of 3D-shape bases, pre-trained using 3D mocap data\cite{cmu_mocap}, is used to learn a sparse representation of human 3D-pose by optimising for reprojection error. It was extended by \cite{zhou2017sparse} via convex relaxation to address bad initialisation in this scheme. Anatomical constraints to regularize the predicted poses w.r.t. limb lengths were introduced in \cite{wang2014robust}. Further use of anatomical constraints in the form of joint-angle-limits and learned pose priors was proposed in \cite{akhter2015pose} to extend \cite{varunECCV2012}. In \cite{Moreno-Noguer_2017_CVPR}, Euclidean inter-joint distance matrix was used to represent 2D and 3D poses with multi-dimensional scaling to obtain 3D-pose from the predicted 3D distance matrix. Some approaches, \cite{bogo2016keep}, estimate the 3D-pose and shape by fitting a 3D statistical model \cite{loper2015smpl} to 2D-pose and leverage inter-penetration constraints. Different from all the previous approaches we employ CVAE to implicitly learn the anatomical constraints and sample 3D-pose candidates. 
 
The method in \cite{kanazawa2018end}, builds upon the framework of \cite{bogo2016keep} to describe a model that estimates the shape, underlying 3D-pose and camera parameters using a re-projection and adversarial loss, which can be trained with 2D-pose datasets and unpaired MoCap datasets. In \cite{martinez2017simple}, a baseline model is proposed that uses a simple fully connected linear network for this task which surprisingly outperforms past approaches. Unlike these discriminative approaches that predict only one 3D-pose from a given 2D-pose, we generate a diverse sample set of 3D-poses.

\myparagraph{Hypothesis Generation} Some previous approaches sample multiple 3D-poses via heuristics. The work in \cite{li2015maximum}, finds the nearest neighbors in a learned latent embedding of human images to estimate the 3D-pose. The approaches in \cite{lee2004proposal} and \cite{sminchisescu2003kinematic}, enumerate 3D-poses using "kinematic-flipping" of the 3D joints, for estimation and tracking, respectively. The Bayesian framework from \cite{simo2013joint} employs a latent-variable generative model with a set of HOG-based 2D part detectors and performs inference using evolutionary algorithms. More recently, \cite{Chen_2017_CVPR} retrieves 3D-pose using nearest neighbor search. \cite{Jahangiri:ICCV2017} uses the pose prior model of \cite{akhter2015pose} to generate multiple hypothesis from a seed 3D-pose, while \cite{wan2017deepskeleton} use "skeleton maps" at different scales to regress 3D-pose hypothesis. Unlike the previous methods, our CVAE based generative model implicitly learns an anatomically consistent pose prior \emph{conditioned} on the input 2D-pose. It affords efficient sampling of a set of candidate 3D-poses without requiring expensive MCMC or graphical model inference or an existing MoCap library. Also, it doesn't need additional image features or structural cues.
Closest to our approach are prior arts that employ generative models for hand-pose estimation. In \cite{spurr2019cross}, \textbf{one-to-one correspondence} is assumed between hand-pose samples in different modalities--RGB, Depth, 2D-pose \& 3D-pose--and a joint latent space is learned via multi-modal VAE. Unfortunately, this assumption between 2D-and-3D poses ignores the inherent ambiguity in 2D-to-3D lifting, while, we explicitly tackle it via CVAE-based probabilistic framework. The work in \cite{disconet2016} generates multiple hand-poses from depth-map to address the prediction uncertainty due to occlusions/missing-values in the input depth-map and uses Maximum-Expected-Utility (MEU) to obtain a pointwise prediction from the generated samples. We use CVAE for generation and employ geometry-inspired ordinal scoring to score and merge multiple samples. \cite{crossingNet2017} learns a probabilistic mapping from depth-map to 3D-pose, to exploit unlabeled data, which is not provably ill-posed. We, however, employ CVAE inspired probabilistic framework to tackle the provable ill-posed nature of 2D-to-3D pose lifting.

\myparagraph{Ordinal Relations} Ordinal relations have previously been explored to estimate depth \cite{zoran2015learning,chen2016single} and reflectance \cite{zhou2015learning,narihira2015learning}. Recently, \cite{pavlakos2018ordinal} and \cite{ronchi2018s} used 2D datasets with ordinal annotations as weak supervision for monocular 3D-pose estimation by imposing a penalty for violation of ordinal depth constraints. Our ordinal prediction network is similar in spirit to \cite{pons2014posebits} that uses a Structural-SVM conditioned on HOG features to predict pose-bits that capture qualitative attributes to facilitate 3D-pose prediction and image retrieval. Unlike \cite{pons2014posebits}, we leverage deep-networks to jointly predict the 2D-pose and depth-ordinal, and generate a diverse sample set of 3D-poses. Concurrent with our work, \cite{wang2018drpose3d} also predict depth ranking and regress 3D-pose from 2D-pose with depth rankings in a coarse-to-fine network.
We differ in the formulation of predicting ordinals as spatial maps, which co-locate with the 2D-pose.

\section{Proposed Approach}

In this Section, we describe the proposed approach. Sec.~\ref{subsection:2DPoseNet} discusses \emph{2DPoseNet} to obtain 2D-pose from an input RGB image followed by Sec.~\ref{subsection:MultiPoseNet} that describes our novel \emph{MultiPoseNet}, for generating multiple 3D-pose samples conditioned on the estimated 2D-pose. In Sec.~\ref{subsection:OrdinalNet}, we discuss \emph{OrdinalNet} to obtain joint-ordinal relations from the image and the estimated 2D-pose. Finally, Sec.~\ref{subsection:OrdinalScore} and ~\ref{subsection:Oracle} describe our strategies for predicting the final 3D-pose from the generated samples : \textbf{(a)} by scoring the generated sample set using ordinal relations, referred to as OrdinalScore, and \textbf{(b)} by using supervision from an Oracle with access to the ground truth 3D-pose, referred to as \emph{OracleScore}. 
\begin{figure}[t]
	\centering
	\includegraphics[width = \linewidth]{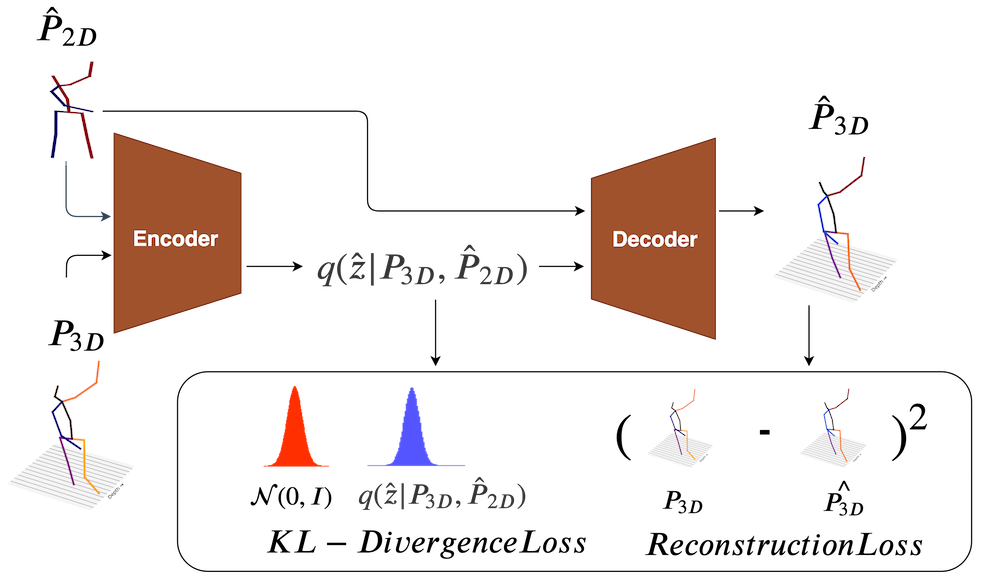} 
    \caption{ MultiPoseNet architecture in training. Note: in GSNN, we sample $z\sim\mathcal{N}(0,I)$ and only need the Decoder.}
    \label{fig:cvae}
    \vspace{-1em}
\end{figure}
\subsection{2DPoseNet: 2D-Pose from Image}
\label{subsection:2DPoseNet}
We use the Stacked Hourglass Model \cite{NewellYD16} with two stacks, as our backbone $C$. The \emph{2DPoseNet} head applies a 1x1 convolution to the intermediate feature representations to regress per-joint heatmaps (Gaussian bumps at target location), from which the predicted 2D pose in pixel coordinates, $\hat{P}_{2D}$, is obtained using Argmax operator.

\subsection{MultiPoseNet: Multiple 3D-Poses from 2D}
\label{subsection:MultiPoseNet}
Recently, Variational Auto-encoders and Generative Adversarial Networks have demonstrated tremendous success in density estimation and synthetic sample generation. Specifically, CVAEs can generate realistic samples conditioned on input variables which is well suited for multi-modal regression mappings \cite{sohn2015learning}. Therefore, we extend the \emph{Baseline} regression model from \cite{martinez2017simple} into a CVAE to tackle the inherent multi-modality of the 2D-to-3D pose mapping and sample an accurate and diverse 3D-pose candidate set $\mathcal{S} = \{\hat{P}^k_{3D}: k \in \{1,2,\ldots K\}\}$ conditioned on the estimated 2D-pose $\hat{P}_{2D}$. We observe that $\mathcal{S}$ has diverse anatomically plausible samples and contains a close match to the actual ground-truth, $P_{3D}$. The detailed architecture for \emph{MultiPoseNet} is depicted in Figure~\ref{fig:cvae}.

\par \myparagraph{Training} The 3D-pose generating CVAE~\cite{sohn2015learning} consists of 
\begin{itemize}
    \item Recognition Network, or Encoder : $Enc(P_{3D},\hat{P}_{2D})$, which operates on an input 3D-pose $P_{3D}$ and a condition $\hat{P}_{2D}$ to output the mean and diagonal covariance for the posterior ${q(\hat{z}|P_{3D},\hat{P}_{2D})}$.
    \vspace{-0.75em}
    \item Decoder : $Dec(\hat{z},\hat{P}_{2D})$, which reconstructs the ground truth $P_{3D}$ by taking as input a latent $\hat{z}$ sampled from the posterior ${q(\hat{z}|P_{3D},\hat{P}_{2D})}$ and the condition 2D-pose $\hat{P}_{2D}$.
\end{itemize}
During training, we optimize the following: 
\begin{align}
\label{eq:cvae}
\mathcal{L}_{CVAE} & = \lambda_1 KL(q(\hat{z}|P_{3D}, \hat{P}_{2D})||p(z|\hat{P}_{2D}))  \\ 
& + \lambda_2\mathbb{E}_{z\sim q(\hat{z}|P_{3D},\hat{P}_{2D})}{||P_{3D} - Dec(\hat{z},\hat{P}_{2D})||}^{2}_{2}, \nonumber
\end{align}
where the prior distribution $p(z|\hat{P}_{2D}))$ is assumed to be $\mathcal{N}(0,I)$, and $KL(x||y)$ is the Kullback-Leibler divergence with $\lambda$s used as hyper-parameters to weight the losses. The expectation in the second term for the reconstruction loss is taken over $K_{train}$ number of samples.

At inference time, the Encoder network is discarded, and $z$ is drawn from the prior $p(z)\sim\mathcal{N}(0,I)$, which introduces inconsistency between the prediction and training pipelines. To remedy this, we set the Encoder equal to the prior network $p(z)\sim\mathcal{N}(0,I)$, that leads to the Gaussian Stochastic Neural Network framework, or GSNN, proposed in \cite{sohn2015learning}. Combining the two we get a hybrid training objective, weighted with $\alpha$:
\begin{align}
    \mathcal{L}_{GSNN} &= \mathbb{E}_{z\sim{N}(0,1)}{||P_{3D} - Dec(z,\hat{P}_{2D})||}^{2}_{2} \\
    \mathcal{L}_{hybrid} &= \alpha{L}_{CVAE} +({1-\alpha}){L}_{GSNN}, 
\end{align}
\myparagraph{Inference} We sample $z\sim\mathcal{N}(0,1)$, and feed ($z$, $\hat{P}_{2D}$) to the Decoder, to obtain $\mathcal{S}_{test} =$ \{$\hat{P}^k_{3D}$: $k \in \{1,2,\ldots,K_{test}\}$\}. 

\subsection{OrdinalNet: Image to Joint-Ordinal Relations}
\label{subsection:OrdinalNet}
The backbone architecture for \emph{OrdinalNet} is same as our \emph{2DPoseNet} i.e. $C$. In order to obtain joint-ordinal relations, we augment $C$ with two additional hourglass stacks. For each human-body joint location $j \in \{1, 2, \ldots, N \}$, three ordinal maps ( $\hat{OM_{1j}}$, $\hat{OM_{2j}}$, and $\hat{OM_{3j}}$ ) are predicted to capture the \emph{lesser than},  \emph{greater than} and \emph{equal} depth relations between joint $j$ and all other joints $i \in \{1, 2, \ldots, N\} $. The ground-truth ordinal maps are generated so that for each joint $j$ there is a Gaussian peak for joint $i \in \{1, 2, \ldots, N\} $ in $one$ of the three ordinal maps ( $OM_{1j}$, $OM_{2j}$, and $OM_{3j}$ ), depending on the depth relation between joint $i$  and joint $j$. We combine the intermediate feature representations and 2D-pose heatmaps from backbone $C$ and \emph{2DPoseNet} as the input, and use L2 loss over predicted ordinal maps, for training our \emph{OrdinalNet}. 
We post-process our estimated ordinal relations via non-maximal suppression on the predicted ordinal maps and associate each peak to its nearest joint-location, which are finally converted into a $16\times16$ joint-ordinal relation matrix $\hat{M}$. The relation between depths $D_{i}, D_{j}$ of joints $i,j \in \{1, 2, \ldots, N \}$ and ground-truth matrix $M$ is:
\[ \hat{M_{ij}} = \left\{
  \begin{array}{lr}
    1 & : D_{i} - D_{j} > 0 \\
    2 & : D_{i} - D_{j} < 0 \\
    3 & : D_{i} - D_{j} \approx 0
  \end{array}
\right.
\]


\subsection{OrdinalScore: Scoring and Aggregating Generated 3D samples}
\label{subsection:OrdinalScore}
So far we have generated a diverse set of estimated 3D-poses from $\hat{P}_{2D}$ only. Next, we seek motivation from the fact that under orthogonal camera projection with constant bone length constraint 2D-pose and joint-ordinal relations between keypoints can ‘almost‘ resolve the true 3D-pose \cite{Taylor:2000:RAO:364058.364079}. The estimated ordinal matrix $\hat{M}$ is used to assign scores to each of the samples $\hat{P^k_{3D}} \in \mathcal{S}$ by the scoring function:
\begin{equation}
 	\label{eq:ord_scoring}
    f(\hat{P}^k_{3D}) =
	\sum\limits_{i, j}\mathbbm{1}{( \hat{M_{ij}}==g(\hat{P}^k_{3D})_{ij} })
\end{equation}
where   $\mathbbm{1}$(condition) is an indicator function,  where $g(\hat{P}^k_{3D})$ is the function that computes the $16\times16$ ordinal matrix for a given  3D-pose and $g(\hat{P}^k_{3D})_{ij}$ represents the ordinal relation of joint i and j. 

 The set of scores for the sampled 3D-poses obtained from an image, $\mathcal{F} = \{f(\hat{P}^k_{3D}): k \in\{1, 2, \dots |\mathcal{S}|\}\}$, is passed through a Softmax operator parameterized by temperature $T$ to obtain a probability distribution function, $p(\hat{P}^k_{3D})=e^{Tf(\hat{P}^k_{3D})}/\sum_{k}{e^{Tf(\hat{P}^k_{3D})}}$. The final output $\hat{P}_{3D}$ is computed as the expectation over the candidates-
\begin{equation}
	\label{eq:ord_aggregation}
    \hat{P}_{3D} = \sum_{k}^{|\mathcal{S}|}{p(\hat{P}^k_{3D}). \hat{P}^k_{3D}} 
\end{equation}
The temperature-based Softmax affords a fine control over the contribution strength of high-score samples vs. the low-scoring samples towards the final aggregation, which makes it robust to noisy pose candidates with respect to the predicted ordinal matrix $\hat{M}$.

\subsection{Supervision from an Oracle}
\label{subsection:Oracle}
The upper-bound accuracy for our approach is given by choosing the closest sample, $\hat{P}^{oracle}_{3D}$, to the ground-truth, $P_{3D}$, from $\mathcal{S}$ using an Oracle that has access to $P_{3D}$.
\begin{equation}
	\label{eq:oracle}
    \hat{P}^{oracle}_{3D} = \argmin_{s \in \mathcal{S}} \|P_{3D} - s\|_2 
\end{equation}

\section{Experiments}
This section discusses the empirical evaluation of the proposed approach. First, we describe the benchmarks that we employed for quantitative evaluation, and provide some important implementation details of our approach. Then, we present quantitative results and compare our method with the state-of-the-art, and provide ablation studies to analyze the performance of our generative model.

\subsection{Datasets}
We make use of the following datasets for training various modules of our pipeline :

\par \myparagraph{CMU Mocap} motion capture dataset consists of diverse 3D-poses with 144 different subjects performing different actions.
We obtain 2D projections from the 3D skeletons using virtual cameras from multiple views, with assumed intrinsic parameters. We employ the obtained 2D-to-3D pose data to train \emph{MultiPoseNet} and the \emph{Baseline} model from \cite{martinez2017simple} for experiments under \emph{unpaired} setting, while \emph{2DPoseNet} and \emph{OrdinalNet} are trained on Human3.6M. Therefore, effectively we train our networks without using any image-to-3D ground-truth data.  

\par \myparagraph{Human3.6M} dataset consists of 3.6 million 3D-poses. It consists of videos and MoCap data of 5 female and 6 male subjects, captured from 4 different viewpoints while they are performing common activities (talking on the phone, walking, greeting, eating, etc.). 

\myparagraph{HumanEva-I} is a small dataset containing 3 subjects (S1, S2, S3) with 3 camera views and fewer actions than Human3.6M. This is a standard dataset for 3D-pose estimation used for benchmarking in previous works.

\subsection{Implementation Details}

\myparagraph{Data Pre-processing:} We take a tight $224\times224$ crop around the person in the input RGB image, $I$, using ground-truth bounding boxes. Following \cite{martinez2017simple}, we process the 3D-poses in camera coordinates and apply standard normalization to the 2D-pose inputs and 3D-pose outputs by subtracting the mean and dividing by the standard deviation, and zero-center the 3D-pose around the hip joint. The 2D-pose contains N=16, and the 3D-pose contains N=17 and N=16 joints for Human3.6M and HumanEva-I respectively.

\myparagraph{2DPoseNet:} We use publicly available Stacked-Hourglass pretrained on MPII \cite{andriluka20142d} as backbone $C$ and \emph{2DPoseNet}, and finetune on Human3.6M and HumanEva-I, following \cite{martinez2017simple}.

\myparagraph{MultiPoseNet:} Its architecture is based on the \emph{Baseline} model in \cite{martinez2017simple} (details in supplementary material). At training time, the expectation in Eq.\ref{eq:cvae} is estimated using $K_{train}=10$ samples. $\lambda_1, \lambda_2$ and $\alpha$ are set to 10, 100, and 0.5 respectively. The network is trained for 200 epochs using Adam \cite{kingma2014adam}, starting with a learning rate of 2.5e-4 with exponential decay and mini-batches size of 256.  At test time, we generate $K_{test}=200$ 3D-pose candidates to get a diverse sample set $\mathcal{S}$. \emph{MultiPoseNet} takes 10 hours to train on a Titan 1080ti GPU.  

\myparagraph{OrdinalNet:} We freeze the weights of our backbone $C$ and \emph{2DPoseNet} after fine-tuning, and train the \emph{OrdinalNet} module using ground-truth ordinal maps for 60 epochs with standard L2 Loss. \emph{OrdinalNet} takes 12 hours to train, on a Titan 1080ti GPU.

\myparagraph{OrdinalScore} The temperature, $T$, is obtained using cross-validation and set to 0.9 for ground truth ordinals, and 0.3 for predicted ordinals. In practice, \emph{OrdinalNet} can sometimes predict contradictory relations i.e $\hat{M}_{ij} \neq \hat{M}_{ji}, \hat{M}_{ii} \neq 3$; we resolve it by setting the diagonal entries of $\hat{M}$ to 3 and mask out elements where $\hat{M}_{ij} \neq \hat{M}_{ji}$ during scoring. Note that for Human3.6M, the ordinal relations w.r.t the extra joint in the 3D-pose are not taken into account by the scoring function in Eq.\ref{eq:ord_scoring}.

\myparagraph{Runtime Details} The run-time for different modules of our pipeline are - \emph{OrdinalNet:} 20ms/image, \emph{MultiPoseNet:} 0.5ms/sample, we take 200 samples/image for inference. The entire pipeline runs at ~10 fps on a commodity graphics card, which is slightly worse than other real-time methods.
\vspace{2mm}

{
\setlength{\tabcolsep}{4pt}
\renewcommand{\arraystretch}{1.1} 
\begin{table*}[ht]
 \centering
 \resizebox{\linewidth}{!}{
   \begin{tabular}{l l c c c c c c c c c c c c c c c c } \\
     & {Protocol 1} & {Direct.} & {Discuss} & {Eating} & {Greet} & {Phone} & {Photo} & {Pose} & {Purch.} & {Sitting} & {SitingD} & {Smoke} & {Wait} & {WalkD} & {Walk} & {WalkT} & {Avg} \\
     \Xhline{4\arrayrulewidth}
     \multirow{15}{*}{PAIR} 

     & Pavlakos {\it et al.}~\cite{Pavlakos_2017_CVPR} &67.4 &71.9 &	66.7	&	69.1 &	72.0	&	77.0	&	65.0&	68.3&	83.7&	96.5 &	71.7	&	65.8 &	74.9 &	59.1 &	63.2 	&	71.9\\
     & Zhou {\it et al.}~\cite{Zhou_2017_ICCV} & 54.82 &60.70 &58.22 &71.4 	& 62.0 & {65.5} &53.8 	&55.6 &75.2 &111.6 &64.1 &66.0 &51.4 &63.2 &55.3 &64.9	\\    
     & Martinez {\it et al.}~\cite{martinez2017simple} &  51.8   &56.2&58.1&59.0&69.5&78.4&55.2&58.1&74.0&94.6&62.3&59.1&65.1&49.5&52.4&62.9 \\ 

  & Sun	{\it et al.}~\cite{DBLP:conf/iccv/0001SLW17} &  52.8	&54.8	& { 54.2}	& {54.3} & {61.8}	& 67.2	& 53.1 & {53.6}	&71.7	&86.7	&61.5	& 	\	{53.4}&61.6	&47.1	&53.4 &59.1 \\  
  
    & Fang {\it et al.}~\cite{DBLP:conf/aaai/FangXWLZ18}	&50.1 & {54.3}	&57.0	&57.1 &66.6	&73.3	&53.4	&55.7	&72.8	&88.6 &60.3	&57.7	&62.7	&47.5	&50.6	&60.4 \\ 
    \cline{2-18}
    & *Pavlakos {\it et al.}~\cite{pavlakos2018ordinal}	 
    &48.5 &54.4 &54.4 &52.0 &59.4 &65.3 &49.9 &52.9 &65.8 &71.1 &56.6 &52.9 &60.9 &44.7 &47.8 &56.2 \\
    
    & **Hossain {\it et al.}-\cite{hossain2018exploiting} & 44.2 & 46.7 & 52.3 & 49.3 & 59.9 & 59.4 & 47.5 & 46.2 & 59.9 & 65.6 & 55.8 & 50.4 & 52.3 & 43.5 & 45.1 & 51.9 \\
    & **Dabral {\it et al.}-\cite{dabral2018learning} & 44.8 & 50.4 & 44.7 & 49.0 & 52.9 & 61.4 & 43.5 & 45.5 & 63.1 & 87.3 & 51.7 & 48.5 & 37.6 & 52.2 & 41.9 & 52.1 \\
    & ***Sun {\it et al.}~\cite{sun2018integral} &47.5 &47.7 &49.5 &50.2 &51.4 &43.8 &46.4 &58.9 &65.7 &49.4 &55.8 &47.8 &38.9 &49.0 &43.8 &49.6 \\
    \cline{2-18}
      & \textbf{Ours (\emph{PRED Ordinals})} & {48.6} & $54.5$ & 54.2 & $55.7$ & $62.6$ & $72.0$ & {50.5 }  & $54.3$ & ${70.0}$ & $ {78.3}$ & ${58.1}$ & $55.4$ &		${61.4}$ & ${45.2}$ &	${49.7}$ & ${58.0}$ \\ 
     \cline{2-18}
       & Ours (\emph{GT Ordinals}) &  $42.9$ & $48.1$ & $47.8$ & $50.2$ & $56.1$ & $65.0$ & $44.9$  & $48.6$ & $61.8$ & $69.9$ & $52.6$ & $50.4$ & $56.0$ & $42.1$ & $45.1$ & $52.1$ \\ 
       & Ours (\emph{Oracle}) &  $37.8$ & $43.2$ & $43.0$ & $44.3$ & $51.1$ & $57.0$ & $39.7$  & $43.0$ & $56.3$ & $64.0$ & $48.1$ & $45.4$ & $50.4$ & $37.9$ & $39.9$ & $46.8$ \\
	\Xhline{4\arrayrulewidth}
     \multirow{4}{*}{UNPAIR} 
     & Martinez {\it et al.}~\cite{martinez2017simple} &  $109.9$ & $112$ & $103.8$ & $115.3$ & $119.3$ & $119.3$ & $114$  & $116.6$ & $118.9$ & $127.3$ & $112.2$ & $119.8$ & $113.4$ & $119.8$ & $111.9$ & $115.6$ \\
     \cline{2-18}
     & \textbf{Ours (\emph{PRED Ordinals})} &  $99.9$ & ${102.7}$ & ${97.9}$ & ${105.9}$ & ${112.0}$ & ${111.7}$ & ${103.9}$ & ${109.4}$ & ${111.7}$ & ${119.4}$ & ${104.8}$ & ${110.8}$ & ${103.2}$ & ${106.9}$ & ${102.3}$ & ${106.8}$ \\  
     \cline{2-18}
     & Ours (\emph{GT Ordinals}) &  $97.9$ & $100.5$ & $95.4$ & $103.7$ & $109.4$ & $108.5$ & $102.0$  & $108.0$ & $107.9$ & $115.4$ & $102.2$ & $108.9$ & $100.8$ & $105.8$ & $100.8$ & $104.4$ \\ 
     & Ours (\emph{Oracle}) & $92.6$ & $94.6$ & $90.6$ & $98.4$ & $103.8$ & $103.3.6$ & $96.6$ & $101.8$ & $101.7$ & $108.8$ & $96.6$ & $102.7$ & $95.3$ & $100.6$ & $96.1$ & $98.9$ \\
\end{tabular}
   }
\vspace{-1em}
\caption{Detailed results on Human3.6M under Protocol 1(no rigid alignment in post-processing). Error is in millimeters(mm). Top: Paired methods (PAIR), Bottom: unpaired methods (UNPAIR). Results for \cite{martinez2017simple} in the unpaired setting were obtained using their publicly available code. * - use additional ordinal training data from MPII and LSP. ** - use temporal information. *** - use soft-argmax for end-to-end training. These strategies are complementary with our approach. }
\label{tab:h36m_prot1}
\end{table*}
}

{
 \vspace{-1em}
\setlength{\tabcolsep}{4pt}
\renewcommand{\arraystretch}{1.1} 
\begin{table*}[ht]
 \centering
 \resizebox{\linewidth}{!}{
   \begin{tabular}{l l c c c c c c c c c c c c c c c c } \\
     & {Protocol 2} & {Direct.} & {Discuss} & {Eating} & {Greet} & {Phone} & {Photo} & {Pose} & {Purch.} & {Sitting} & {SitingD} & {Smoke} & {Wait} & {WalkD} & {Walk} & {WalkT} & {Avg} \\
     \Xhline{4\arrayrulewidth}
     \multirow{9}{*}{PAIR} 
 & Zhou {\it et al.}~\cite{Zhou_2017_ICCV} &  47.9 & 48.8 & 52.7 & 55.0 & 56.8 &49.0 & 45.5 & 60.8 & 81.1 & 53.7  & 65.5 & 51.6 & 50.4 & 54.8 &55.9  & 55.3 \\    
     & Pavlakos {\it et al.}~\cite{Pavlakos_2017_CVPR}  &   47.5 & 50.5 & 48.3 & 49.3 & 50.7 & 55.2 & 46.1 & 48.0 & 61.1 & 78.1 & 51.1 & 48.3 & 52.9 & 41.5 & 46.4 &51.9 \\
       & Martinez {\it et al.}~\cite{martinez2017simple} & 39.5 & 43.2 & 46.4 & 47.0 & 51.0  &56.0 & 41.4 & 40.6 & 56.5 & 69.4 & 49.2 & 45.0 & 49.5 & 38.0 & 43.1 &47.7 \\  
     & Fang {\it et al.}~\cite{DBLP:conf/aaai/FangXWLZ18} &	38.2 &	 41.7 & 43.8 & 44.9 & 48.5 & 55.3 & 40.2 & 38.2 & 54.5& 64.4 & 47.2 & 44.3& 47.3& 36.7& 41.7& 45.7\\
     & Sun {\it et al.}~\cite{DBLP:conf/iccv/0001SLW17} & 42.1 & 44.3 & 45.0 &45.4& 51.5& 53.0 &43.2 &41.3& 59.3 & 73.3 & 51.0 & 44.0 &  48.0 &38.3& 44.8& 48.3\\
     \cline{2-18}
      & *Pavlakos {\it et al.}~\cite{pavlakos2018ordinal} & 34.7 & 39.8 & 41.8 & 38.6 & 42.5 & 47.5 & 38.0 & 36.6 & 50.7 & 56.8 & 42.6 & 39.6 & 43.9 & 32.1 & 36.5 & 41.8\\
      & **Hossain {\it et al.}-\cite{hossain2018exploiting} & 36.9 & 37.9 & 42.8 & 40.3 & 46.8 &  46.7  & 37.7 & 36.5 &  48.9 & 52.6 &  45.6  & 39.6 & 43.5  & 35.2  & 38.5 & 42.0 \\
      & **Dabral {\it et al.}-\cite{dabral2018learning} & 28.0 & 30.7 & 39.1 & 34.4 & 37.1 & 44.8 & 28.9 & 31.2 & 39.3 & 60.6 & 39.3 & 31.1 & 25.3 & 37.8 & 28.4 & 36.3 \\
      & ***Sun {\it et al.}~\cite{sun2018integral} & - & - & - & - & - & - & - & - & - & - & - & - & - & - & - & 40.6 \\
      \cline{2-18}
  & \textbf{Ours (\emph{PRED Ordinals})} &  $35.3$ & $35.9$ & $45.8$ & $42.0$ & $40.9$ & $52.6$ &  $36.9$  & $35.8$ & $43.5$ & $51.9$ & $44.3$ & $38.8$ & $45.5$ & $29.4$ & $34.3$ & $40.9$ \\ 
       \cline{2-18}
       & Ours (\emph{GT Ordinals}) &  $31.3$ & $31.0$ & $39.3$ & $37.0$ & 37.2 & 47.8	& 32.5 & 32.1 & 	39.8 & 47.3	& 40.0 & 34.7 & 41.8 & 27.5 & $31.0$ & $36.7$ \\ 
	   & Ours (\emph{Oracle}) & 27.6 & 27.5 & 34.9 & 32.3 & 33.3 & 42.7 & 28.7 & 28.0 & 36.1 &	42.7 & 36.0 & 30.7 & 37.6 & 24.3 & 27.1 & 32.7 \\
	\Xhline{4\arrayrulewidth}
     \multirow{4}{*}{UNPAIR} 
     & Martinez {\it et al.}~\cite{martinez2017simple} & 62.6 & 64.3 & 62.5 & 67.4 & 72.2 & 70.8 & 64.9 & 61.2 & 82.1 & 92.4 & 76.8 & 66.7 & 71.7 & 79.5 & 73.1 & 71.3 \\
     \cline{2-18}
     & \textbf{Ours(\emph{PRED Ordinals})} & 62.9 & 65.6 & 61.8 & 67.1 & 72.2 & 69.3 & 65.6 & 63.8 & 81.3 & 91.0 & 74.5 & 66.5 & 70.8 & 74.7 & 70.9 & 70.5 \\
    \cline{2-18}
    & Ours(\emph{GT Ordinals}) &  62.9 & 65.3 & 60.7 & 66.9 & 71.3 & 68.4 & 65.2 & 63.2 & 80.1 & 89.3 & 73.5 & 66.1 & 70.5 & 74.7 & 70.9 & 70.0 \\  
    & Ours (\emph{Oracle}) & 56.8 & 59.2 & 55.0 & 59.6 & 65.6 & 62.0 & 58.4 & 56.5 & 74.2 & 82.8 & 67.6 & 60.0 & 63.6 & 68.2 & 64.3 & 63.6\\
\end{tabular}
   }
\vspace{-1em}
\caption{Detailed results on Human3.6M under Protocol 2(rigid alignment in post-processing). Top: Paired methods (PAIR), Bottom: unpaired methods (UNPAIR). Results for \cite{martinez2017simple} in the unpaired setting were obtained using their publicly available code.}
\label{tab:h36m_prot2}
 \vspace{-1em}
\end{table*}
}

\subsection{Quantitative Evaluation}
In this sub-section, we report the results of our model and compare it against the prior state-of-the-art on Human3.6M and HumanEva-I dataset. We report three evaluation metrics to demonstrate the benefits of our approach: 
    \par \emph{PRED Ordinals}: Uses the OrdinalScore strategy with the ordinal relations predicted by \emph{OrdinalNet}.
    \par \emph{GT Ordinals}: Uses the OrdinalScore strategy with the ground truth ordinal relations.
    \par \emph{Oracle}: Uses the Oracle for final prediction, which gives the best results.
\subsubsection{Evaluation on Human3.6M} Following the literature, we use two standard protocols to train and evaluate our results.
\myparagraph{Protocol-1}: The training set consists of 5 subjects (S1, S5, S6, S7, S8),
while the test set includes 2 subjects (S9, S11). The original 50FPS frame rate is down-sampled to 10 FPS and the evaluation is carried out on sequences coming from all 4 cameras and all trials. The reported error metric is Mean Per Joint Position Error (MPJPE) i.e. the Euclidean distance from the estimated 3D-pose, $\hat{P}_{3D}$, to the ground-truth, $P_{3D}$, averaged over 17 joints of the Human3.6M skeletal model. 
\myparagraph{Protocol-2}: Subjects S1, S5, S6, S7, S8 and S9 are used for training and S11 for testing. The error metric used is Procrustes Aligned MPJPE (PA MPJPE) which is the MPJPE calculated after rigidly aligning the predicted pose with the ground-truth. 
\par Table~\ref{tab:h36m_prot1} and Table~\ref{tab:h36m_prot2} show our results for Protocol-1 and Protocol-2, respectively. In the paired setting, we train each module, that is, \emph{2DPoseNet}, \emph{OrdinalNet} and \emph{MultiPoseNet}, using paired image-to-3D pose annotations from Human3.6M. Under this setting, we achieve competitive results using \emph{PRED Ordinals} for scoring. The use of \emph{GT Ordinals} takes us close to the state-of-the-art. We are worse only to the methods that either use additional ordinal training data \cite{pavlakos2018ordinal}, temporal information \cite{dabral2018learning, hossain2018exploiting} and/or soft-argmax \cite{sun2018integral} (denoted by *s), all of which is compatible with our approach and is expected to improve the performance further. Finally, we outperform all existing methods using \emph{Oracle} supervision. Although it's an unfair comparison, it demonstrates that our CVAE-generated sample set contains candidate poses that are very close to the ground-truth pose, thus validating our sample-generation based approach.     


\par \myparagraph{Without Paired 3D Supervision:} The modular nature of our pipeline allows us to train the 2D-to-3D lifting module on a separate MoCap library that has no intersection with the training images for \emph{2DPoseNet}, \emph{OrdinalNet}. It affords training our pipeline without the costly and laborious acquisition of paired image-to-3D annotations. We demonstrate it by training \emph{MultiPoseNet} on the CMU MoCap dataset, which consists of only 3D MoCap data, and report the results on the test-set of Human3.6M. Note that the MoCap dataset is only needed for training, not for testing. The 3D-poses from CMU MoCap are virtually projected to their corresponding 2D-projections, with the camera at the origin and pelvis at a distance of 5500mm. We have used the intrinsic camera parameters from Human3.6M to bring the distribution of 2D-projections closer to the Human3.6M test set. We also rotate the 3D-poses by 90, 180, and 270 degrees, for data augmentation. The obtained 2D-to-3D pose dataset is used to train the \emph{Baseline} model \cite{martinez2017simple} and \emph{MultiPoseNet}. The estimated 2D-poses and ordinals are obtained from \emph{2DPoseNet} and \emph{OrdinalNet}, both of which are trained on Human3.6M. We emphasize that Human3.6M is only used for learning 2D-pose and ordinal estimation, therefore, we don't use any image-to-3D annotation during training. Since, two different sources are used for the image-to-2D/ordinal and 2D-to-3D modules, we call it \emph{unpaired setting}. The results of these experiments are reported in Table~\ref{tab:h36m_prot1} and \ref{tab:h36m_prot2} in the bottom rows. 

Our \emph{PRED Ordinals} based method outperforms the \emph{Baseline} regression model \cite{martinez2017simple} and with the use of \emph{GT Ordinals} and \emph{Oracle} the performance only increases. It shows that our framework can learn without image-to-3D annotation and is also robust to domain shift.

\subsubsection{Evaluation on HumanEva-I} 
Under the protocol from \cite{Kostrikov2014DepthSR}, we evaluate our model on HumanEva-I. Training uses subjects S1, S2, S3 under different view-points and action-sequences Jogging and Walking, while testing is carried out on the validation sequences for all three subjects as testing data. All the modules are trained using HumanEva-I. The model error is reported as the reconstruction error after rigid transformation. We obtain state-of-the-art results using the \emph{Oracle} estimate and close to state-of-the-art with \emph{PRED Ordinals} and \emph{GT Ordinals} on HumanEva-I, reported in Table~\ref{tab:heva}.   

\begin{table}[h!]
\resizebox{  \linewidth}{!}{
\vspace{-2em}
\begin{tabular}{ c | c c c | c c c | c }
  \multirow{2}{*}{} 
      & \multicolumn{3}{c}{Jogging} 
          & \multicolumn{3}{c}{Walking} \\ \hline
  & S1 & S2 & S3 & S1 & S2 & S3  & Avg\\  \hline
  Kostrikov {\it et al.}~\cite{Kostrikov2014DepthSR} & 44.0 & 30.9 &  41.7 & 57.2  & 35.0 &  33.3 & 40.3 \\      
    Yasin {\it et al.}~\cite{yasin2016dual} & 35.8 & 32.4 & 41.6 & 46.6  & 41.4 &  35.4 & 38.9  \\      
      Moreno-Noguer {\it et al.}~\cite{Moreno-Noguer_2017_CVPR} & 19.7 & 13.0 &  24.9 & 39.7 & 20.0 &  21.0 &  26.9  \\ 
      Pavlakos {\it et al.}~\cite{Pavlakos_2017_CVPR} & 22.1 & 21.9 &29.0 & 29.8 & 23.6 & 26.0  & 25.5 \\ 
        Martinez {\it et al.}~\cite{martinez2017simple} & 19.7 & 17.4 & 46.8 &26.9 &18.2 & 18.6&  24.6  \\  \hline     
  \textbf{Ours (\emph{PRED Ordinals})} & 19.3 & 12.5 & 41.8 & 40.9 & 22.1 & 18.6  & 25.9\\  \hline
  Ours (\emph{GT Ordinals}) & 19.1 & 12.4 & 41.5 & 40.6 & 21.9 & 18.5  & 25.7\\  \hline
  Ours (\emph{Oracle}) & 17.4& 11.0 & 39.5 & 38.5 & 20.1 & 16.7 & 23.9  \\
\end{tabular}
}
\small
\vspace{-1em}
\caption{Results of our model on HumanEva-I dataset and a comparison with previous work. Numbers reported are mean reconstruction error in mm computed after rigid transformation.
}
\label{tab:heva}
\end{table}

\begin{figure*}[ht]
	\centering
    \begin{subfigure}[t]{0.49\textwidth}
        \includegraphics[width=\textwidth, ]{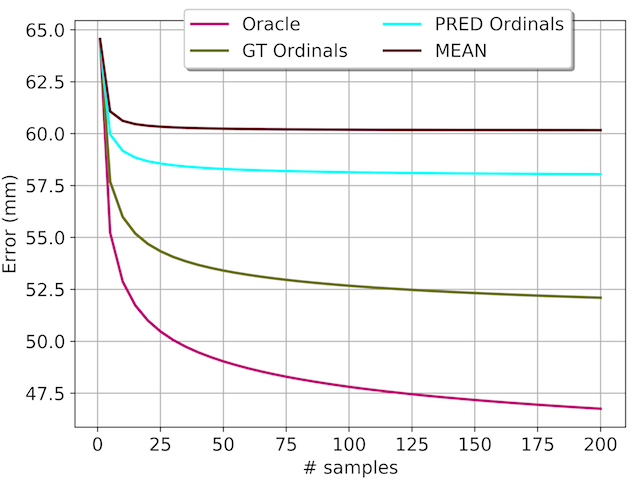}
        \caption{ Oracle vs OrdinalScore vs MEAN }
        \label{fig:Oracle_vs_OrdinalScore}
        \end{subfigure}
     \begin{subfigure}[t]{0.49\textwidth}
        \includegraphics[width=\textwidth, ]{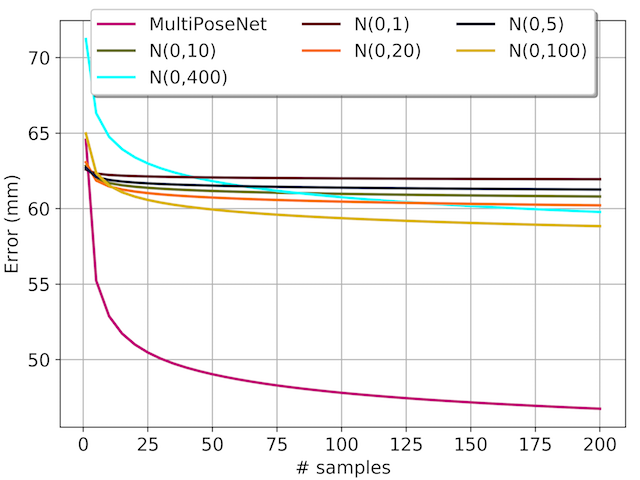} 
        \caption{ MultiPoseNet vs Baseline sampling }
        \label{fig:CVAE_vs_NaiveSampling}
        \end{subfigure}
        \vspace{-3mm}
    	\caption{Ablation studies. (a) Effect of increasing number of samples on Oracle, OrdinalScore and MEAN  estimate (b) Comparison of MultiPoseNet versus Baseline sampling using Oracle supervision.} 
	\label{fig:ablation}
    \vspace{-2mm}
\end{figure*}

\begin{figure*}[ht]
    \begin{subfigure}[t]{0.18\textwidth}
        \includegraphics[width=\textwidth]{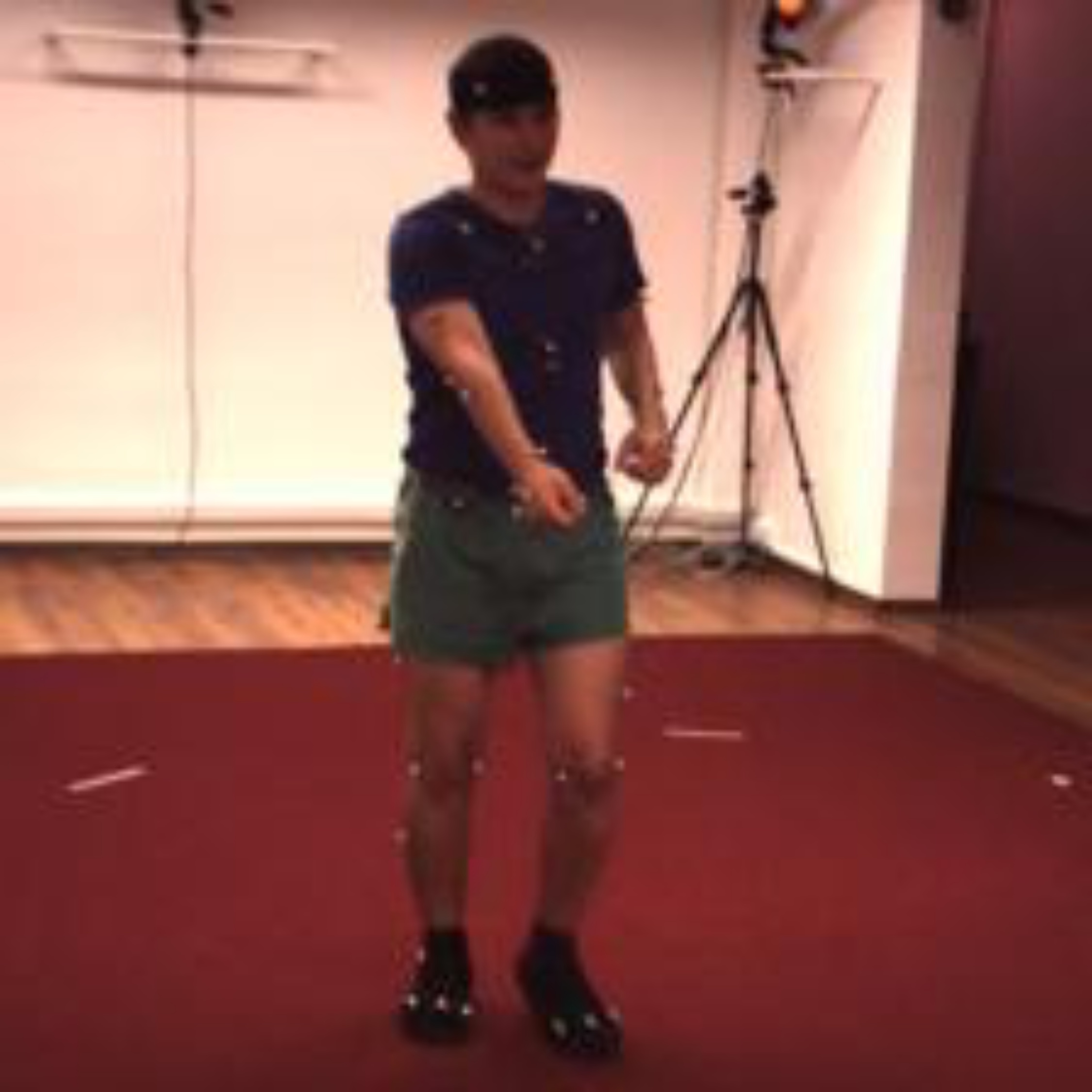}
        \vspace{-2mm}
        \end{subfigure}
    \begin{subfigure}[t]{0.18\textwidth}
        \includegraphics[width=\textwidth]{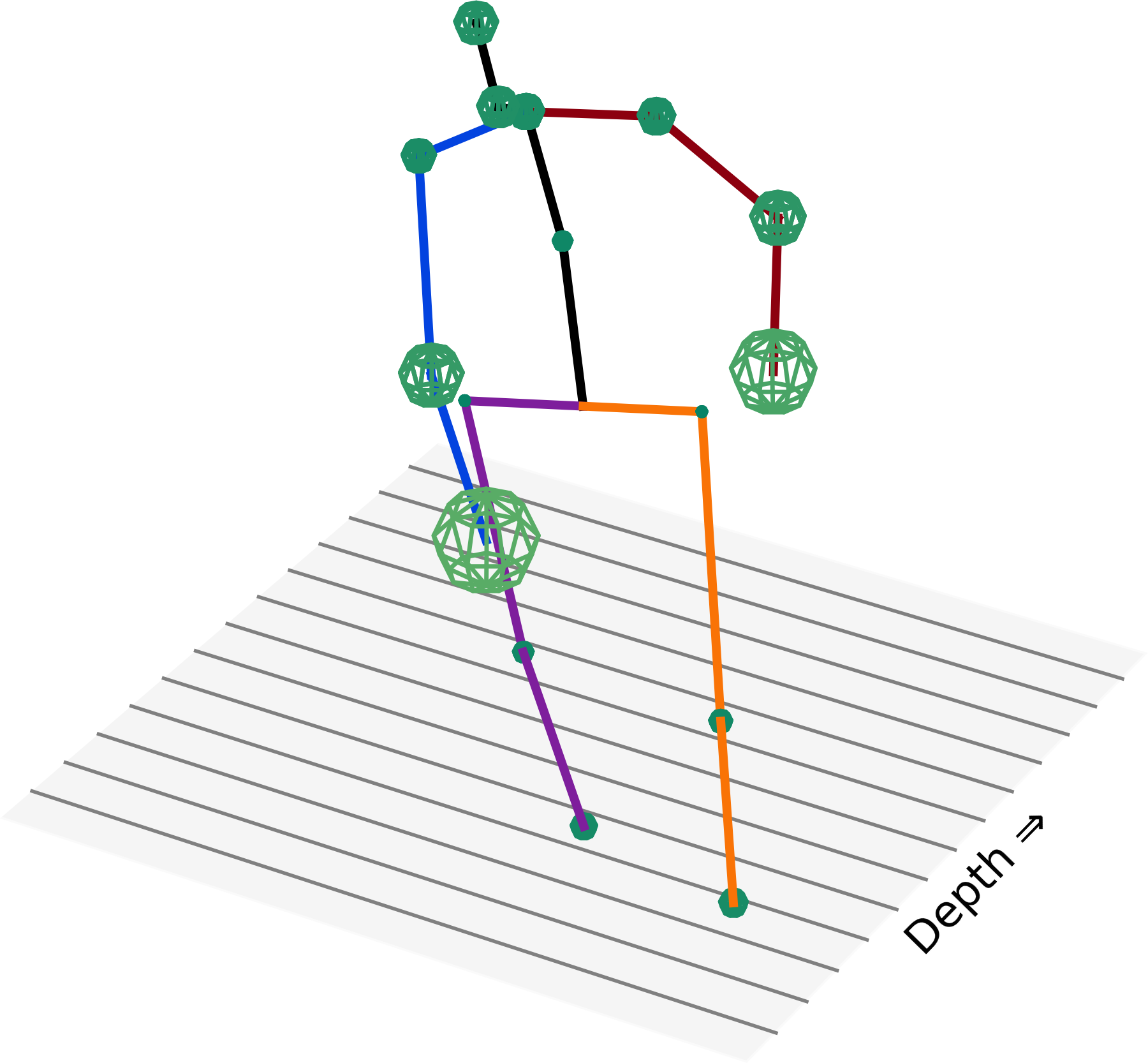}
        \vspace{-2mm}
        \end{subfigure}
     \begin{subfigure}[t]{0.18\textwidth}
        \includegraphics[width=\textwidth]{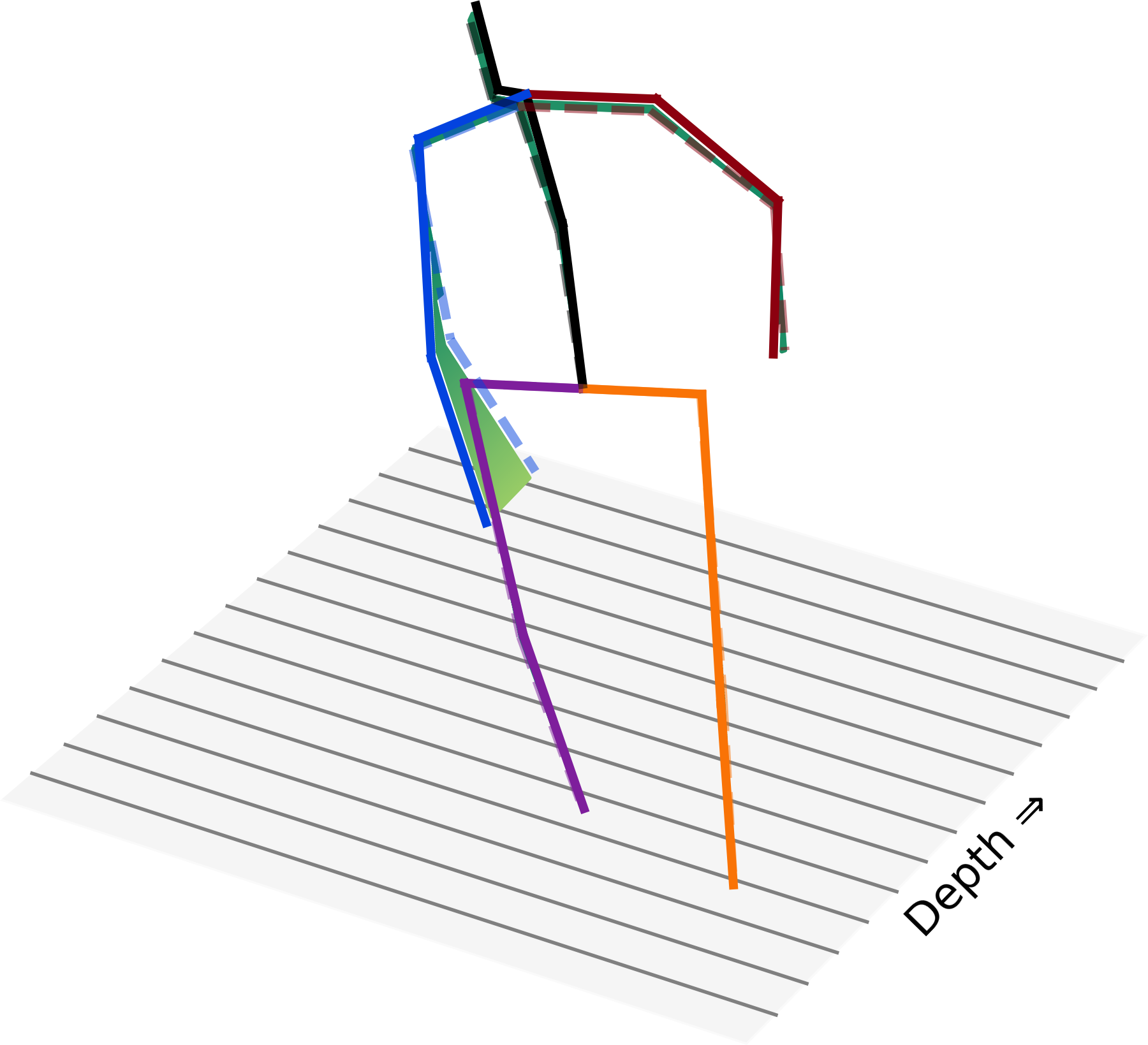}
        \vspace{-2mm}
        \end{subfigure}
    \begin{subfigure}[t]{0.18\textwidth}
        \includegraphics[width=\textwidth]{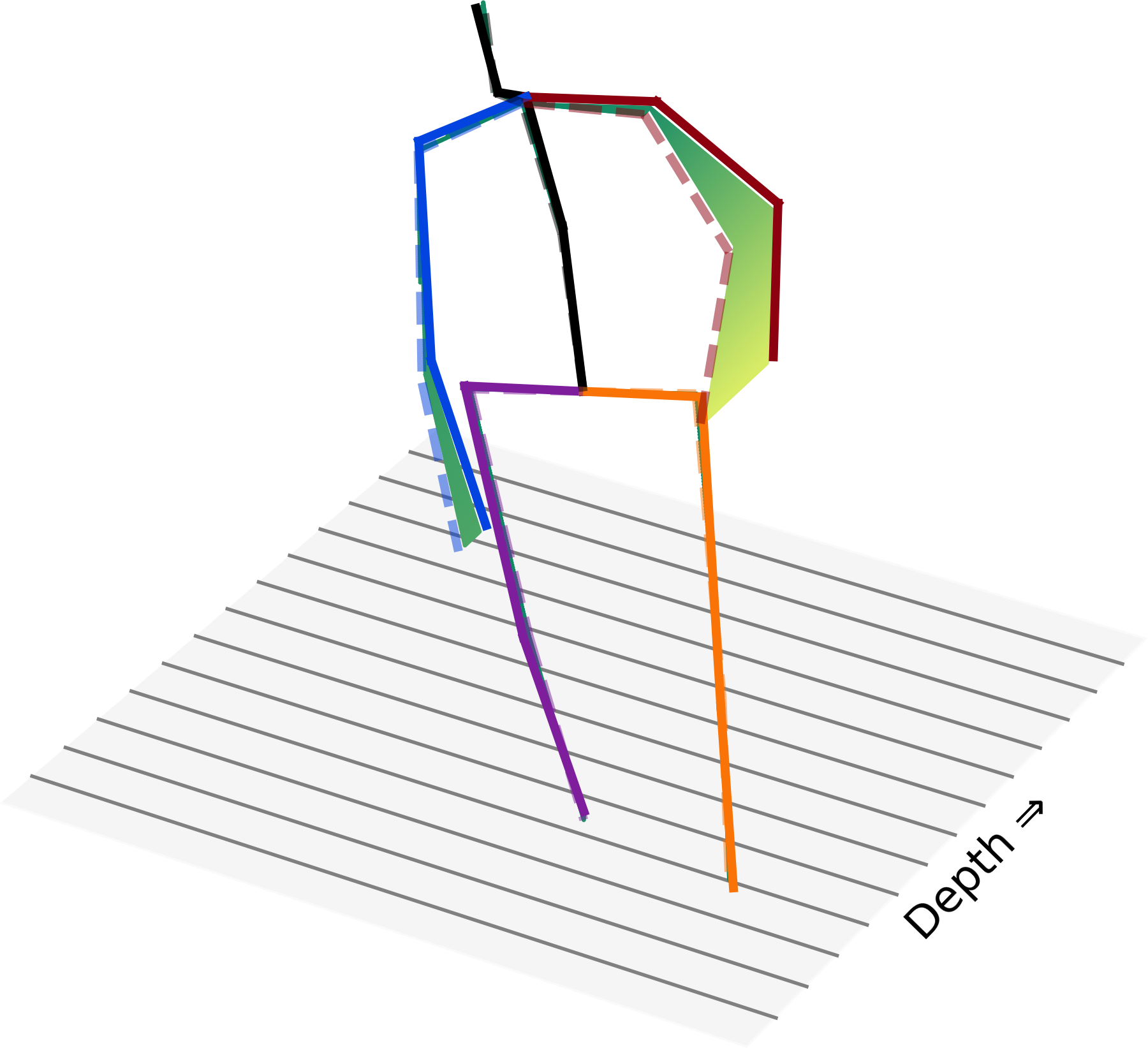}
        \vspace{-2mm}
        \end{subfigure}
    \begin{subfigure}[t]{0.18\textwidth}
        \includegraphics[width=\textwidth]{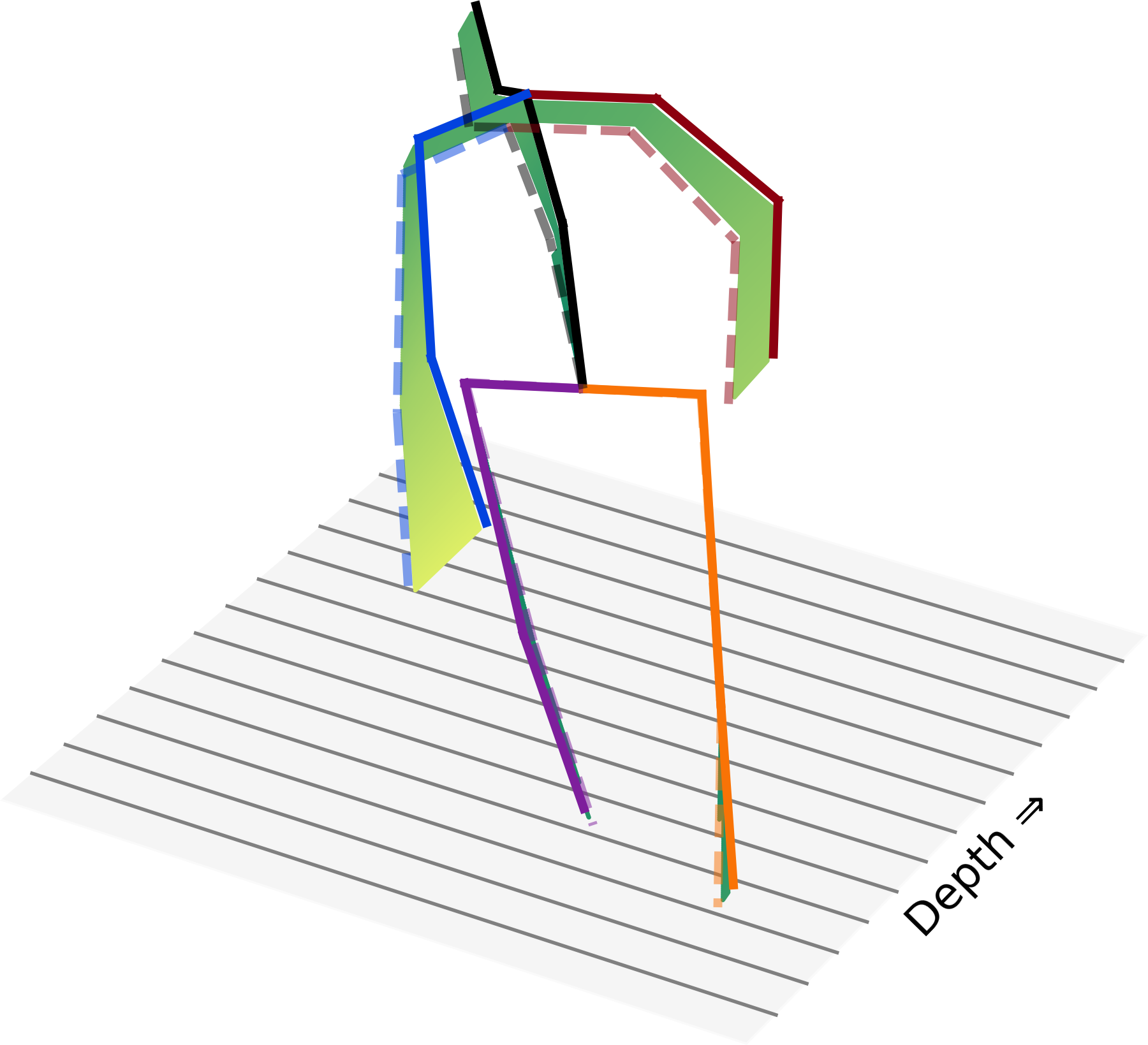}
        \vspace{-2mm}
        \end{subfigure}
    \begin{subfigure}[t]{0.03\textwidth}
        \includegraphics[width=\textwidth, height=85pt]{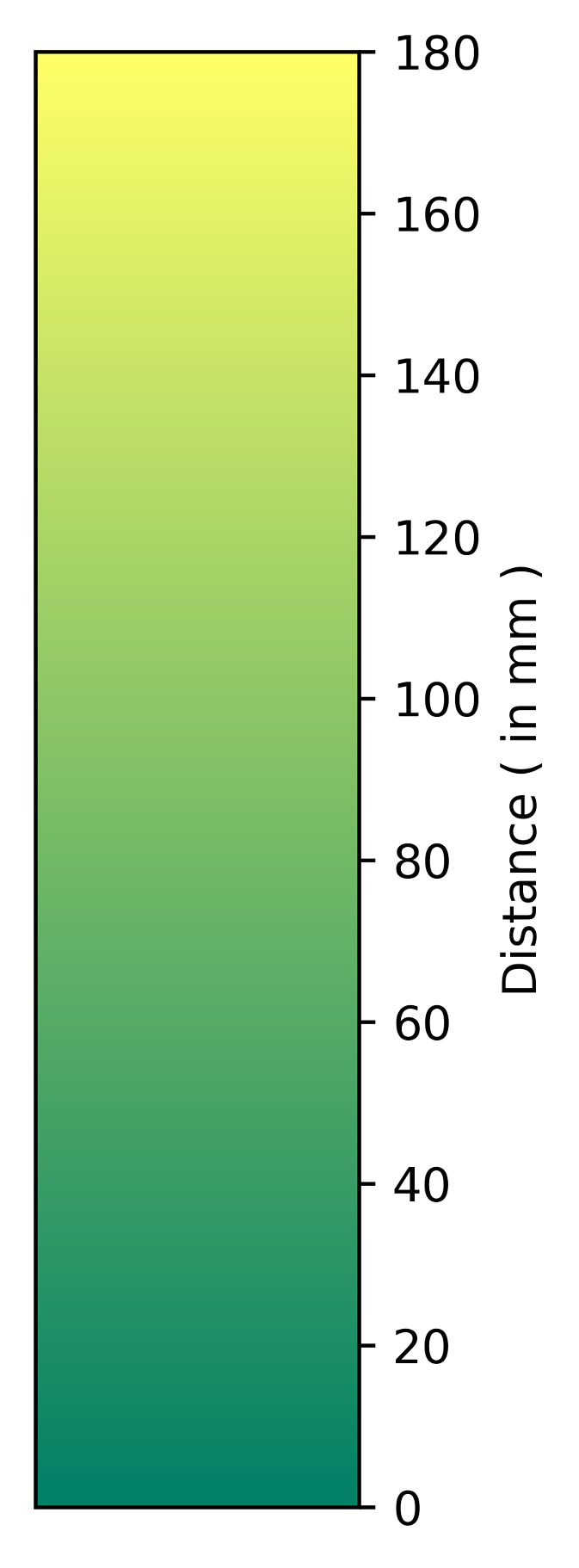}
        \vspace{-2mm}
        \end{subfigure}
    \begin{subfigure}[t]{0.18\textwidth}
        \includegraphics[width=\textwidth]{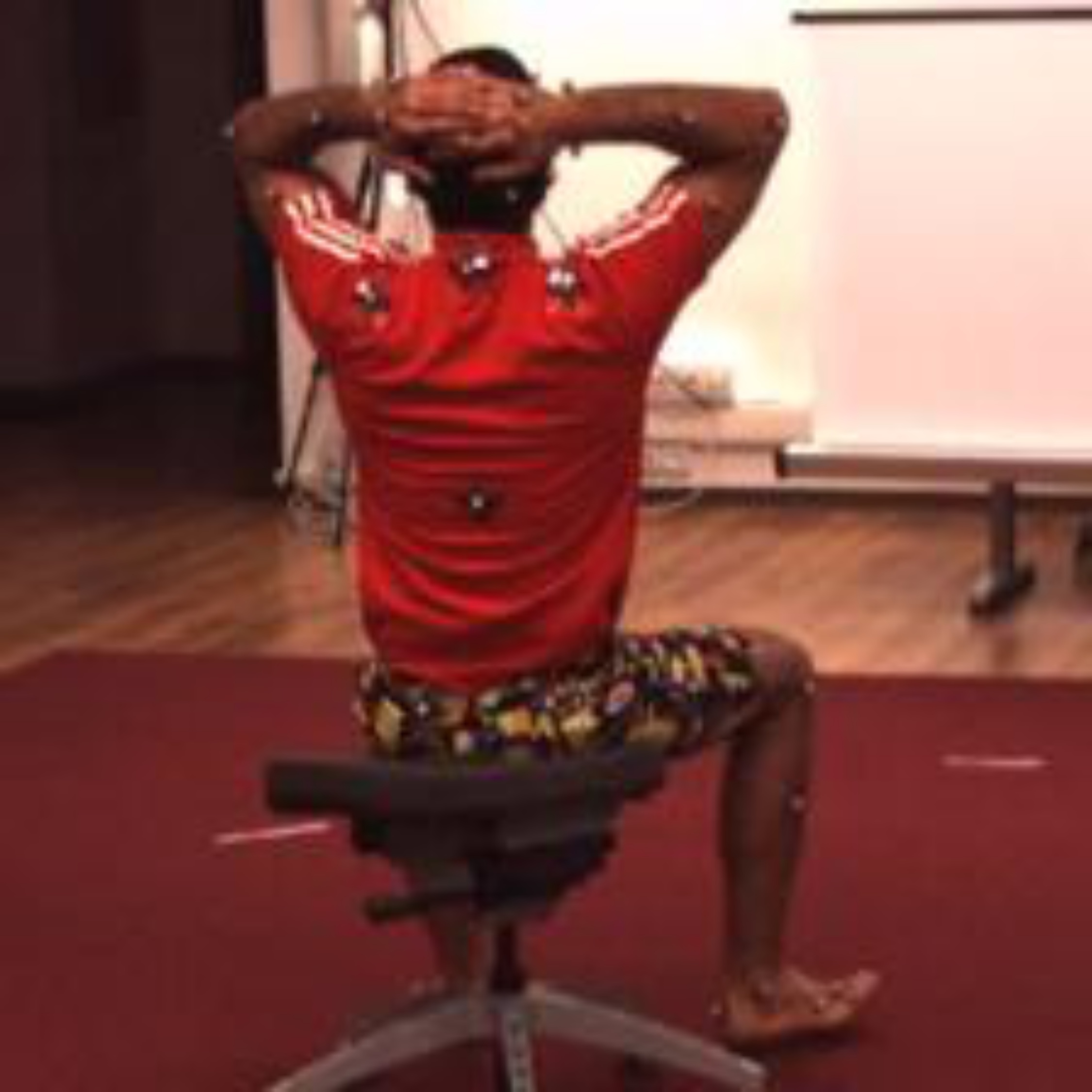}
        \vspace{-2mm}
        \end{subfigure} \quad
    \begin{subfigure}[t]{0.18\textwidth}
        \includegraphics[width=\textwidth]{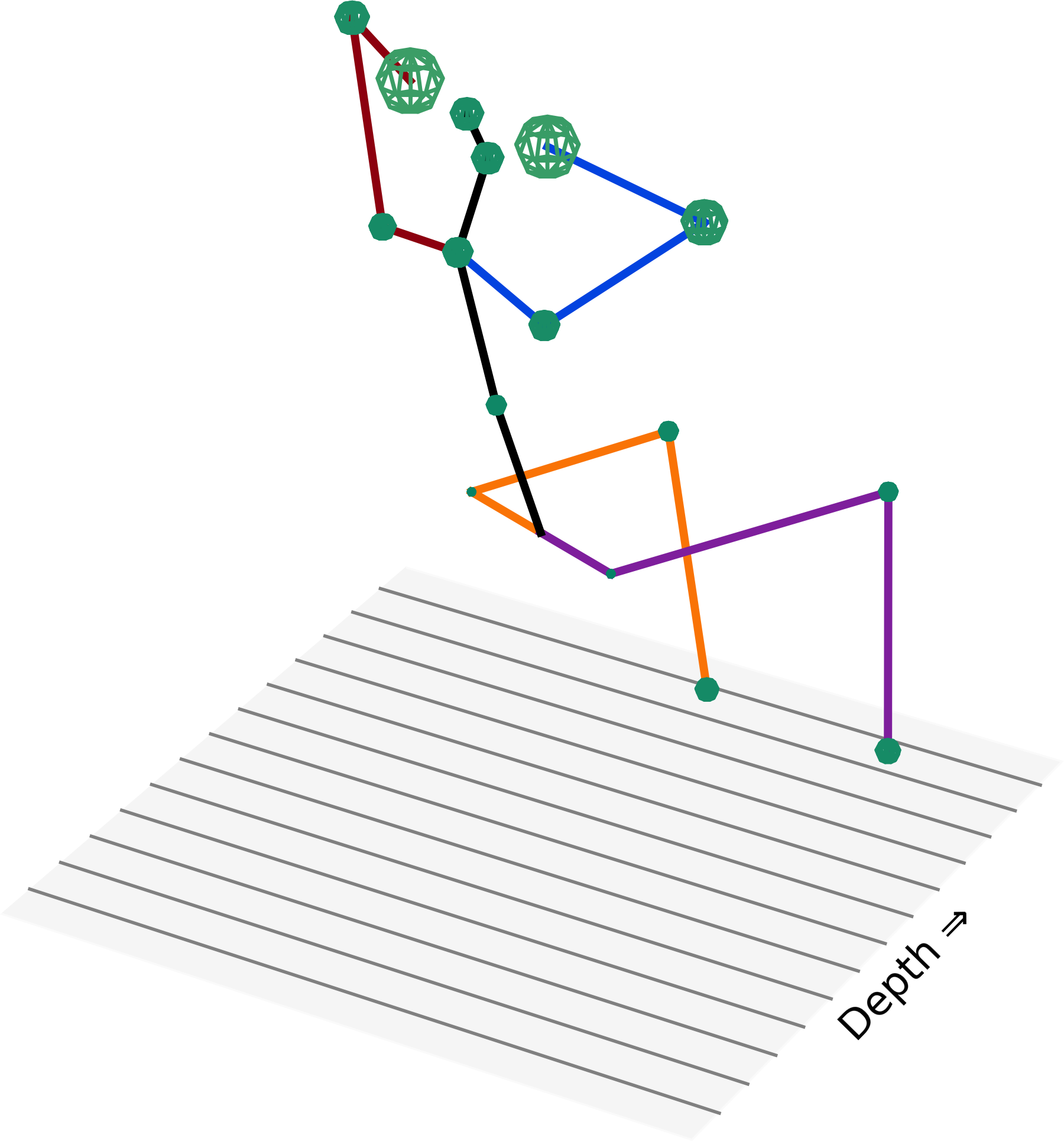}
        \vspace{-2mm}
        \end{subfigure}
     \begin{subfigure}[t]{0.18\textwidth}
        \includegraphics[width=\textwidth]{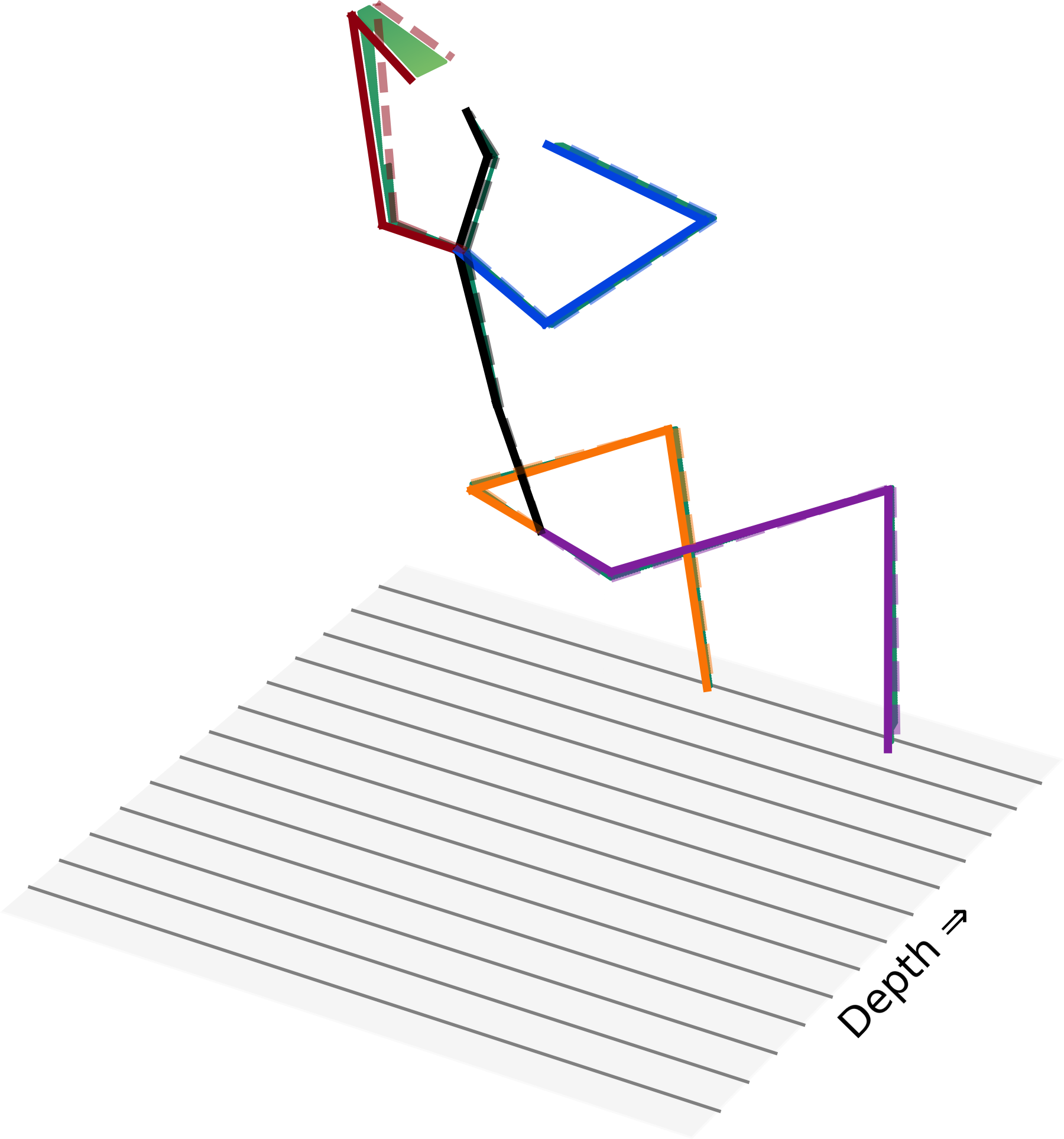}
        \vspace{-2mm}
        \end{subfigure}
    \begin{subfigure}[t]{0.18\textwidth}
        \includegraphics[width=\textwidth]{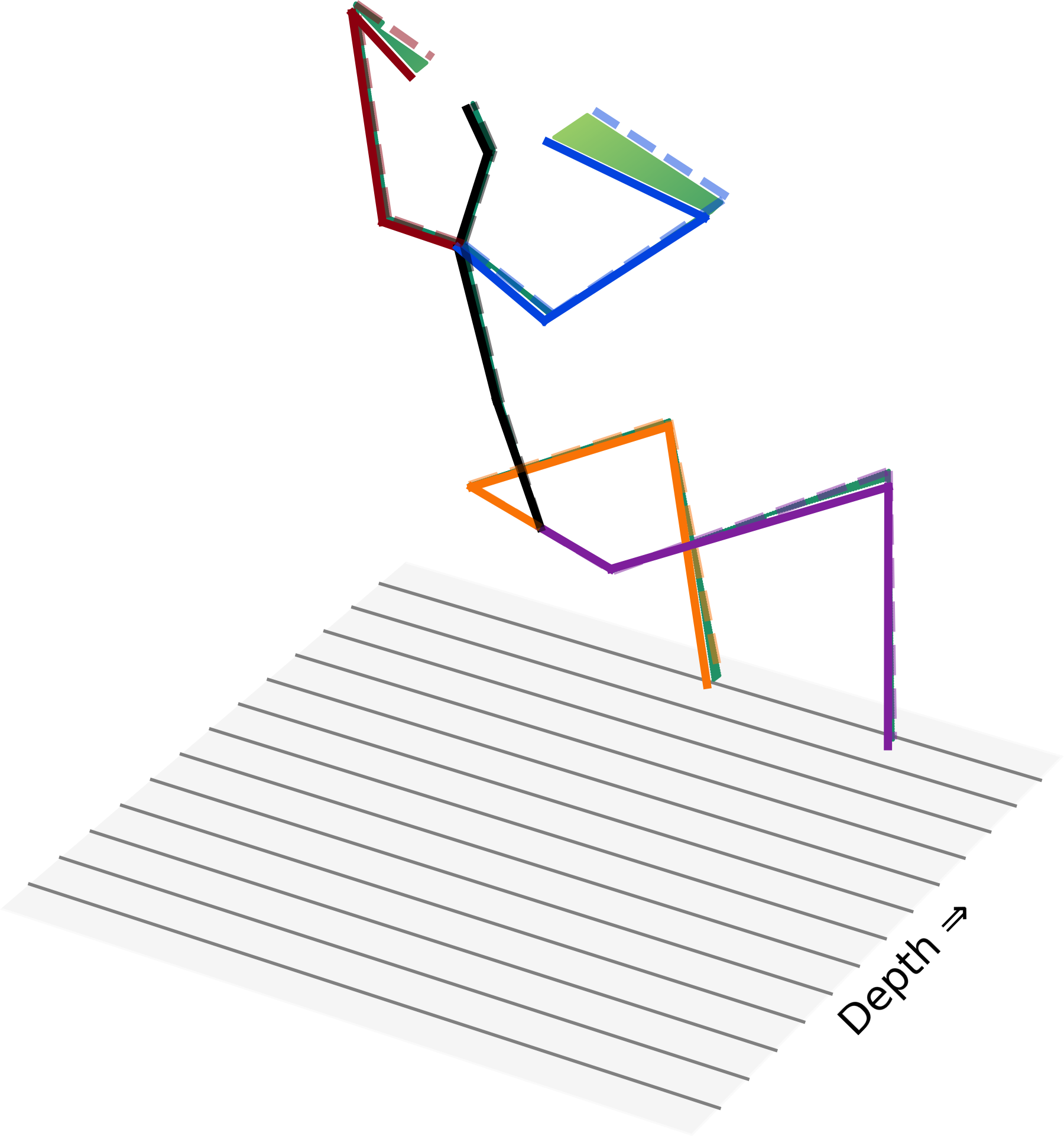}
        \vspace{-2mm}
        \end{subfigure}
    \begin{subfigure}[t]{0.18\textwidth}
        \includegraphics[width=\textwidth]{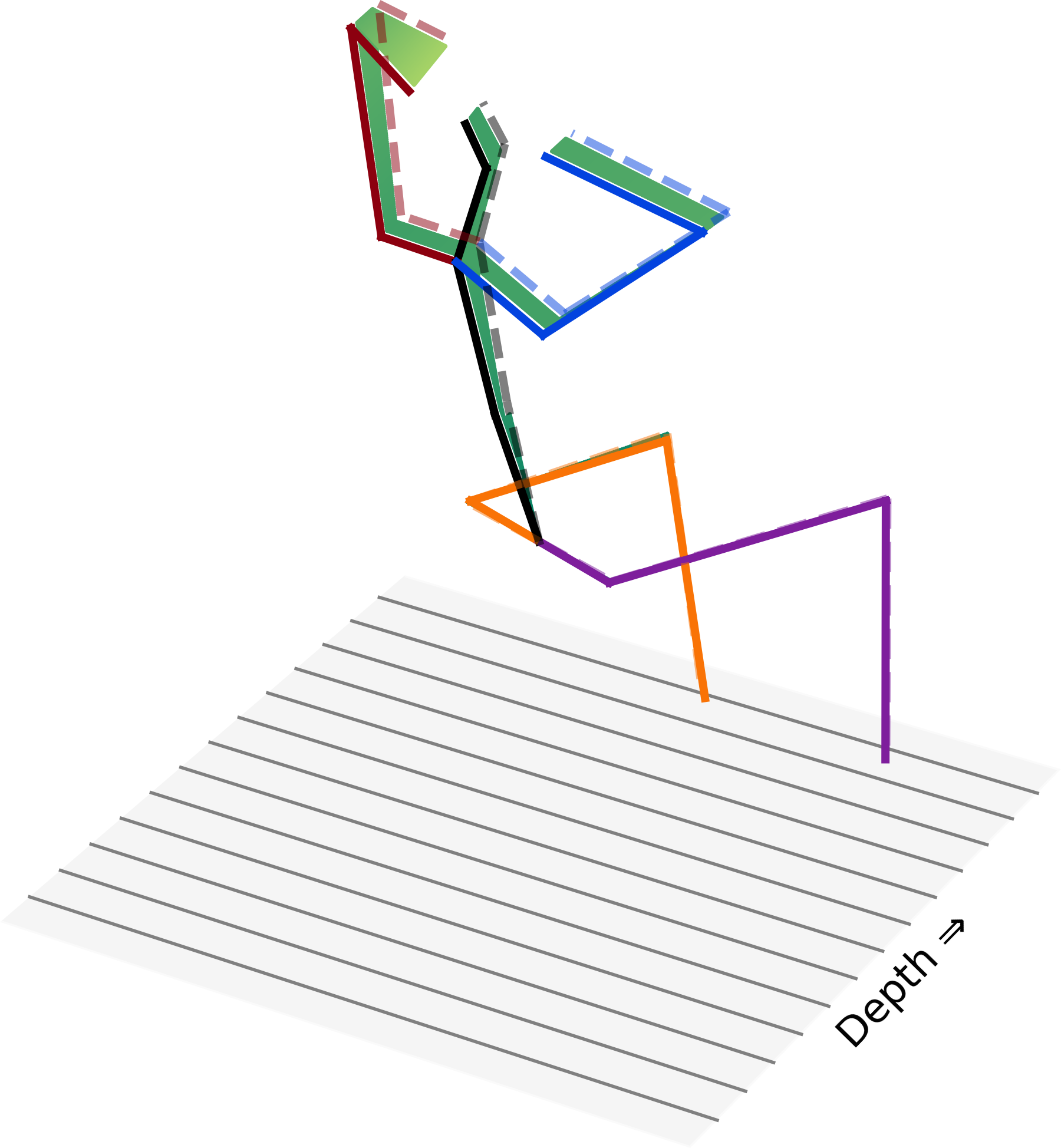}
        \vspace{-2mm}
        \end{subfigure}\quad
     \begin{subfigure}[t]{0.03\textwidth}
        \includegraphics[width=\textwidth, height=85pt]{std_colorbar.png}
        \end{subfigure}
    \vspace{-3mm}
    \caption{Sample diversity on Human3.6M test-set. From L-R: Input Image, MEAN Pose with per-joint standard deviation around each joint, and 3 different SAMPLES overlaid on top of MEAN pose. MEAN is solid and SAMPLE is dashed, with displacement field in between. Note that wrist and elbow show maximum variance. Best viewed in color with zoom.}
	\label{fig:qualitative_viz}
    \vspace{-3mm}
\end{figure*}

\begin{figure*}[ht]
	\centering
    \begin{subfigure}[t]{0.49\textwidth}
        \includegraphics[width=\textwidth, ]{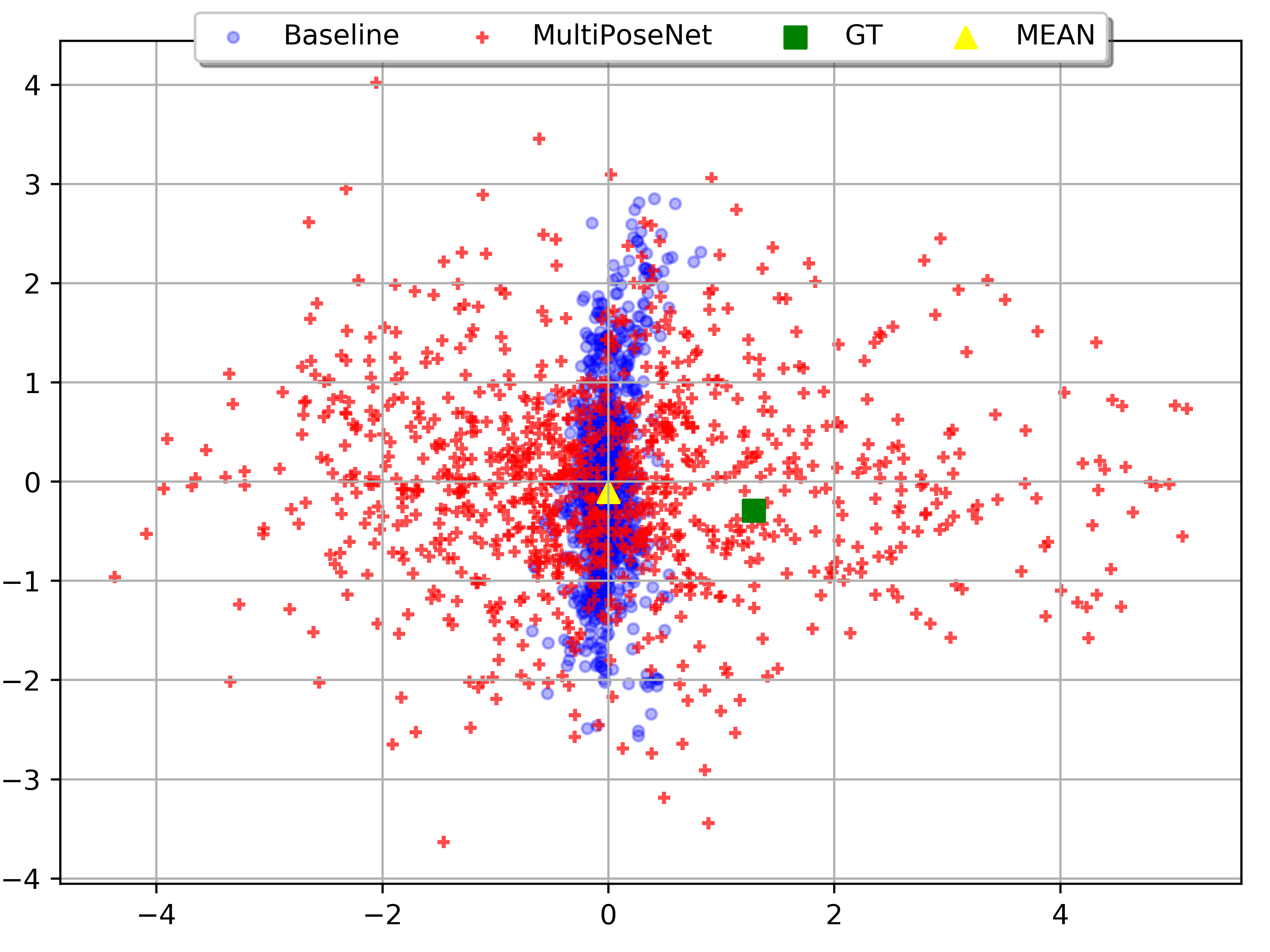}
        \vspace{-2mm}
        \end{subfigure}
     \begin{subfigure}[t]{0.49\textwidth}
        \includegraphics[width=\textwidth, ]{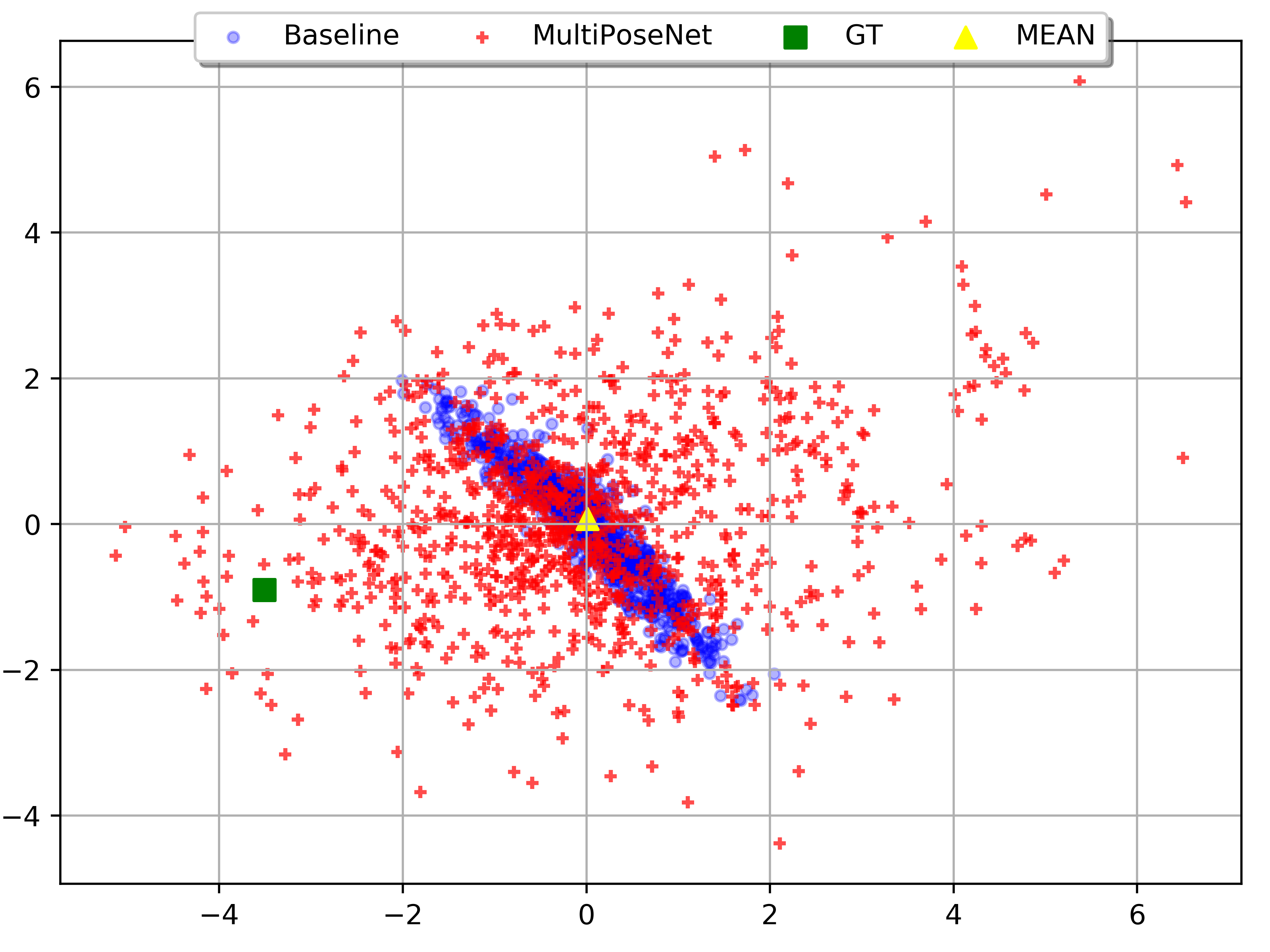} 
        \vspace{-2mm}
        \end{subfigure}
        \vspace{-2mm}
    	\caption{Samples from MultiPoseNet and Baseline ( using a variance of 100 ) mapped to Euclidean space using ISOMAP \cite{tenenbaum2000global}. Note that MultiPoseNet produces much more diverse samples that are likely to be near the GT pose. } 
    \vspace{-3mm}
	\label{fig:sample diversity}
\end{figure*}

\subsection{OrdinalNet Accuracy}
The \emph{OrdinalNet} accuracy is obtained by comparing the ground-truth ordinals, $M$, with the predicted ordinals, $\hat{M}$. The results on the validation set for Human3.6M and HumanEva-I are 86.8\% and 81\% respectively. 

\subsection{Ablation Studies}

\myparagraph{Effect of Increasing Sample Set Size:} In Figure~\ref{fig:Oracle_vs_OrdinalScore}, we plot the value of different error estimates on Protocol-1 of Human3.6 with increasing number of samples. \emph{MEAN} denotes the uniform average of all samples. We observe that the \emph{MEAN} improves with the number of samples, but saturates quickly. The \emph{Oracle} performance keeps on improving with the number of samples, which validates the intuition that the chance of obtaining close to ground-truth pose increases with more samples. Consequently, the estimated 3D-pose, either using \emph{PRED Ordinals} or \emph{GT Ordinals}, keeps improving with more samples, as is evident from their respective curves. This demonstrates that the proposed ordinal scoring is an effective strategy for weighted averaging of the generated samples. 

\myparagraph{Sampling Baseline:} Here, we compare a \emph{Baseline} sampling strategy against our CVAE-based generative sampling. \emph{Baseline} sampling treats each joint-location as independent Gaussian distribution with the mean as the output of the \emph{Baseline} regression model\cite{martinez2017simple} and variance from \{1,5,10,20,100,400\}. Each joint-location is sampled independently to obtain a 3D-pose. \emph{Oracle} supervision is used for both \emph{Baseline} sampling and our \emph{MultiPoseNet} sampling to obtain the final 3D-pose. Figure~\ref{fig:CVAE_vs_NaiveSampling} shows the comparison of \emph{MultiPoseNet} with \emph{Baseline} sampling on Protocol-1 of Human3.6 with increasing number of samples. It's evident that \emph{Baseline} performs poorly and does not improve steeply with more number of samples. It also begins to worsen with higher variance of 400mm as the samples become more absurd. On the other hand, \emph{MultiPoseNet} improves its estimate by close to 20mm and the slope of the curve indicates further potential gains by sampling more. 

\subsection{Sample Diversity}
\myparagraph{Qualitative Analysis:} To assess the feasibility of the proposed approach to generate a diverse set of plausible 3D-pose candidates from a given 2D-pose, we show the \emph{MEAN} pose, per-joint standard deviation, and a few candidate 3D poses for two different images from the Human3.6M test set in Figure~\ref{fig:qualitative_viz}. We observe meaningful variations across different body parts and poses with relatively higher variance around, the hardest to predict, wrist and elbow joints.

\myparagraph{Visualisation Using Dimensionality Reduction:} To visualize the distribution of generated candidate 3D-poses, we map the samples from \emph{MultiPoseNet} and \emph{Baseline} sampling (with a variance of 100) into Euclidean space using Isomap \cite{tenenbaum2000global}. Fig.~\ref{fig:sample diversity} shows 1000 samples using both \emph{MultiPoseNet} and \emph{Baseline} sampling for two different input 2D-poses, along with the ground truth 3D-pose and the \emph{MEAN} estimate of \emph{MultiPoseNet}. Interestingly, the samples from \emph{Baseline} are clustered narrowly around the \emph{MEAN}, whereas \emph{MultiPoseNet} samples are diverse and are more likely to be near the GT 3D-pose. 

\vspace{-0.25em}
\section{Conclusion and Future Work}

This article presented a novel framework for monocular 3D-pose estimation that uses a conditional variational autoencoder for sampling 3D-pose candidates which are scored and weighted-averaged together using ordinal relations, predicted from a deep CNN. The proposed method achieves close to state-of-the-art results on two benchmark datasets using OrdinalScore, and state-of-the-art results using an Oracle with access to the ground truth 3D-pose. The CVAE has been shown to learn a generative model that synthesizes diverse 3D-pose samples consistent with the input 2D-pose, thereby dealing with the ambiguity in lifting from 2D-to-3D. It can also be trained without paired image-to-3D annotations, and still yields competitive results.   

\section{Acknowledgements}
This research was funded in part by Mercedes Benz Research and Development, India. We also thank Bernt Schiele for providing valuable feedback on the manuscript.

{\small
\bibliographystyle{ieee_fullname}
\bibliography{egbib_final}

\begin{thebibliography}{10}\itemsep=-1pt

\bibitem{akhter2015pose}
Ijaz Akhter and Michael~J Black.
\newblock Pose-conditioned joint angle limits for 3d human pose reconstruction.
\newblock In {\em CVPR}, 2015.

\bibitem{andriluka20142d}
Mykhaylo Andriluka, Leonid Pishchulin, Peter Gehler, and Bernt Schiele.
\newblock 2d human pose estimation: New benchmark and state of the art
  analysis.
\newblock In {\em CVPR}, 2014.

\bibitem{bogo2016keep}
Federica Bogo, Angjoo Kanazawa, Christoph Lassner, Peter Gehler, Javier Romero,
  and Michael~J Black.
\newblock Keep it smpl: Automatic estimation of 3d human pose and shape from a
  single image.
\newblock In {\em ECCV}, 2016.

\bibitem{disconet2016}
Diane Bouchacourt, M Pawan~Kumar, and Sebastian Nowozin.
\newblock Disco nets: Dissimilarity coefficient networks.
\newblock In {\em NeurIPS}, 2016.

\bibitem{Chen_2017_CVPR}
Ching-Hang Chen and Deva Ramanan.
\newblock 3d human pose estimation = 2d pose estimation + matching.
\newblock In {\em CVPR}, 2017.

\bibitem{chen2016single}
Weifeng Chen, Zhao Fu, Dawei Yang, and Jia Deng.
\newblock Single-image depth perception in the wild.
\newblock In {\em NIPS}, 2016.

\bibitem{cmu_mocap}
CMU.
\newblock Carnegie mellon university graphics lab - motion capture library,
  2014. http://mocap.cs.cmu.edu/.

\bibitem{dabral2018learning}
Rishabh Dabral, Anurag Mundhada, Uday Kusupati, Safeer Afaque, Abhishek Sharma,
  and Arjun Jain.
\newblock Learning 3d human pose from structure and motion.
\newblock In {\em ECCV}, 2018.

\bibitem{DBLP:conf/aaai/FangXWLZ18}
Haoshu Fang, Yuanlu Xu, Wenguan Wang, Xiaobai Liu, and Song{-}Chun Zhu.
\newblock Learning pose grammar to encode human body configuration for 3d pose
  estimation.
\newblock In {\em AAAI}, 2018.

\bibitem{hossain2018exploiting}
Mir Rayat~Imtiaz Hossain and James~J Little.
\newblock Exploiting temporal information for 3d human pose estimation.
\newblock In {\em ECCV}, 2018.

\bibitem{h36m_pami}
Catalin Ionescu, Dragos Papava, Vlad Olaru, and Cristian Sminchisescu.
\newblock Human3.6m: Large scale datasets and predictive methods for 3d human
  sensing in natural environments.
\newblock {\em PAMI}, 2014.

\bibitem{Jahangiri:ICCV2017}
Ehsan Jahangiri and Alan~L. Yuille.
\newblock Generating multiple diverse hypotheses for human 3d pose consistent
  with 2d joint detections.
\newblock In {\em ICCV}, 2017.

\bibitem{kanazawa2018end}
Angjoo Kanazawa, Michael~J Black, David~W Jacobs, and Jitendra Malik.
\newblock End-to-end recovery of human shape and pose.
\newblock In {\em CVPR}, 2018.

\bibitem{kingma2014adam}
Diederik~P Kingma and Jimmy Ba.
\newblock Adam: A method for stochastic optimization.
\newblock In {\em ICLR}, 2015.

\bibitem{Kostrikov2014DepthSR}
Ilya Kostrikov and Juergen Gall.
\newblock Depth sweep regression forests for estimating 3d human pose from
  images.
\newblock In {\em BMVC}, 2014.

\bibitem{lee2004proposal}
Mun~Wai Lee and Isaac Cohen.
\newblock Proposal maps driven mcmc for estimating human body pose in static
  images.
\newblock In {\em CVPR}, 2004.

\bibitem{li2015maximum}
Sijin Li, Weichen Zhang, and Antoni~B Chan.
\newblock Maximum-margin structured learning with deep networks for 3d human
  pose estimation.
\newblock In {\em ICCV}, 2015.

\bibitem{loper2015smpl}
Matthew Loper, Naureen Mahmood, Javier Romero, Gerard Pons-Moll, and Michael~J
  Black.
\newblock Smpl: A skinned multi-person linear model.
\newblock {\em ACM transactions on graphics (TOG)}, 2015.

\bibitem{martinez2017simple}
Julieta Martinez, Rayat Hossain, Javier Romero, and James~J Little.
\newblock A simple yet effective baseline for 3d human pose estimation.
\newblock In {\em ICCV}, 2017.

\bibitem{Moreno-Noguer_2017_CVPR}
Francesc Moreno-Noguer.
\newblock 3d human pose estimation from a single image via distance matrix
  regression.
\newblock In {\em CVPR}, 2017.

\bibitem{narihira2015learning}
Takuya Narihira, Michael Maire, and Stella~X Yu.
\newblock Learning lightness from human judgement on relative reflectance.
\newblock In {\em CVPR}, 2015.

\bibitem{NewellYD16}
Alejandro Newell, Kaiyu Yang, and Jia Deng.
\newblock Stacked hourglass networks for human pose estimation.
\newblock In {\em ECCV}, 2016.

\bibitem{omran2018neural}
Mohamed Omran, Christoph Lassner, Gerard Pons-Moll, Peter Gehler, and Bernt
  Schiele.
\newblock Neural body fitting: Unifying deep learning and model based human
  pose and shape estimation.
\newblock In {\em 3DV}, 2018.

\bibitem{pavlakos2018ordinal}
Georgios Pavlakos, Xiaowei Zhou, and Kostas Daniilidis.
\newblock Ordinal depth supervision for 3d human pose estimation.
\newblock In {\em CVPR}, 2018.

\bibitem{Pavlakos_2017_CVPR}
Georgios Pavlakos, Xiaowei Zhou, Konstantinos~G. Derpanis, and Kostas
  Daniilidis.
\newblock Coarse-to-fine volumetric prediction for single-image 3d human pose.
\newblock In {\em CVPR}, 2017.

\bibitem{pons2014posebits}
Gerard Pons-Moll, David~J Fleet, and Bodo Rosenhahn.
\newblock Posebits for monocular human pose estimation.
\newblock In {\em CVPR}, 2014.

\bibitem{varunECCV2012}
Varun Ramakrishna, Takeo Kanade, and Yaser Sheikh.
\newblock Reconstructing 3d human pose from 2d image landmarks.
\newblock In {\em ECCV}, 2012.

\bibitem{ronchi2018s}
Matteo~Ruggero Ronchi, Oisin Mac~Aodha, Robert Eng, and Pietro Perona.
\newblock It's all relative: Monocular 3d human pose estimation from weakly
  supervised data.
\newblock {\em BMVC}, 2018.

\bibitem{sigal2010humaneva}
Leonid Sigal, Alexandru~O Balan, and Michael~J Black.
\newblock Humaneva: Synchronized video and motion capture dataset and baseline
  algorithm for evaluation of articulated human motion.
\newblock {\em International journal of computer vision}, 2010.

\bibitem{simo2013joint}
Edgar Simo-Serra, Ariadna Quattoni, Carme Torras, and Francesc Moreno-Noguer.
\newblock A joint model for 2d and 3d pose estimation from a single image.
\newblock In {\em CVPR}, 2013.

\bibitem{sminchisescu2003kinematic}
Cristian Sminchisescu and Bill Triggs.
\newblock Kinematic jump processes for monocular 3d human tracking.
\newblock In {\em CVPR}, 2003.

\bibitem{sohn2015learning}
Kihyuk Sohn, Honglak Lee, and Xinchen Yan.
\newblock Learning structured output representation using deep conditional
  generative models.
\newblock In {\em NIPS}, 2015.

\bibitem{spurr2019cross}
Adrian Spurr, Jie Song, Seonwook Park, and Otmar Hilliges.
\newblock Cross-modal deep variational hand pose estimation.
\newblock In {\em CVPR}, 2018.

\bibitem{DBLP:conf/iccv/0001SLW17}
Xiao Sun, Jiaxiang Shang, Shuang Liang, and Yichen Wei.
\newblock Compositional human pose regression.
\newblock In {\em ICCV}, 2017.

\bibitem{sun2018integral}
Xiao Sun, Bin Xiao, Fangyin Wei, Shuang Liang, and Yichen Wei.
\newblock Integral human pose regression.
\newblock In {\em ECCV}, 2018.

\bibitem{Taylor:2000:RAO:364058.364079}
Camillo~J. Taylor.
\newblock Reconstruction of articulated objects from point correspondences in a
  single uncalibrated image.
\newblock {\em Comput. Vis. Image Underst.}, 80(3):349--363, Dec. 2000.

\bibitem{tenenbaum2000global}
Joshua~B Tenenbaum, Vin De~Silva, and John~C Langford.
\newblock A global geometric framework for nonlinear dimensionality reduction.
\newblock {\em Science}, 2000.

\bibitem{crossingNet2017}
C. {Wan}, T. {Probst}, L.~V. {Gool}, and A. {Yao}.
\newblock Crossing nets: Combining gans and vaes with a shared latent space for
  hand pose estimation.
\newblock In {\em CVPR}, 2017.

\bibitem{wan2017deepskeleton}
Qingfu Wan, Wei Zhang, and Xiangyang Xue.
\newblock Deepskeleton: Skeleton map for 3d human pose regression.
\newblock {\em arXiv preprint arXiv:1711.10796}, 2017.

\bibitem{wang2014robust}
Chunyu Wang, Yizhou Wang, Zhouchen Lin, Alan~L Yuille, and Wen Gao.
\newblock Robust estimation of 3d human poses from a single image.
\newblock In {\em CVPR}, 2014.

\bibitem{wang2018drpose3d}
Min Wang, Xipeng Chen, Wentao Liu, Chen Qian, Liang Lin, and Lizhuang Ma.
\newblock Drpose3d: Depth ranking in 3d human pose estimation.
\newblock {\em IJCAI}, 2018.

\bibitem{wei2016convolutional}
Shih-En Wei, Varun Ramakrishna, Takeo Kanade, and Yaser Sheikh.
\newblock Convolutional pose machines.
\newblock In {\em CVPR}, 2016.

\bibitem{yasin2016dual}
Hashim Yasin, Umar Iqbal, Bjorn Kruger, Andreas Weber, and Juergen Gall.
\newblock A dual-source approach for 3d pose estimation from a single image.
\newblock In {\em CVPR}, 2016.

\bibitem{zhou2015learning}
Tinghui Zhou, Philipp Krahenbuhl, and Alexei~A Efros.
\newblock Learning data-driven reflectance priors for intrinsic image
  decomposition.
\newblock In {\em ICCV}, 2015.

\bibitem{Zhou_2017_ICCV}
Xingyi Zhou, Qixing Huang, Xiao Sun, Xiangyang Xue, and Yichen Wei.
\newblock Towards 3d human pose estimation in the wild: A weakly-supervised
  approach.
\newblock In {\em ICCV}, 2017.

\bibitem{zhou2016sparseness}
Xiaowei Zhou, Menglong Zhu, Kosta Derpanis, and Kostas Daniilidis.
\newblock Sparseness meets deepness: 3d human pose estimation from monocular
  video.
\newblock In {\em CVPR}, 2016.

\bibitem{zhou2017sparse}
Xiaowei Zhou, Menglong Zhu, Spyridon Leonardos, and Kostas Daniilidis.
\newblock Sparse representation for 3d shape estimation: A convex relaxation
  approach.
\newblock {\em PAMI}, 2017.

\bibitem{zoran2015learning}
Daniel Zoran, Phillip Isola, Dilip Krishnan, and William~T Freeman.
\newblock Learning ordinal relationships for mid-level vision.
\newblock In {\em ICCV}, 2015.

\end{thebibliography}
}

\end{document}